# Neural Computation with Rings of Quasiperiodic Oscillators

E. A. Rietman and R. W. Hillis
Physical Sciences Inc., Andover, MA 01810

## Summary

We are proposing an innovative approach to neural computation with applications to adaptive robots. This approach will enable robots to have complex responses to unfamiliar situations without the need for either a computationally intensive central processor or preprogrammed prior anticipation of all possible situations.

Conventional robots achieve adaptive behavior by either digital programmed world-models (Bekey, 2005) or through large numbers of finite state machines programmed for small tasks – sensor input/actuator output (Arkin, 1999). The former approach requires massive amounts of up-front programming and results in a brittle computational system. New and/or unexpected events will result in robot behavior not necessarily appropriate to the situation since the robot can only draw from a limited library of preprogrammed behaviors. The latter approach has the advantage of not requiring a world model but suffers from the same problem of not responding appropriately in many situations. Biological systems do not suffer from these limitations because they are adaptive.

It is important to differentiate learning and adaptation. Learning behavior in biological systems requires producing new synaptic connections. Learning behavior in hardware requires programmable or adjustable resistive connections and has been well explored (Murray and Tarassenko, 1994). Adaptive behavior in biological systems does not require learning. The organism simply changes its behavior for a given situation. For many organisms there is no learning involved in this adaptation – they are wired that way at birth.

This pre-wiring in biological organisms is based on central pattern generators (CPG). Simple organisms have no central processor or brain, but they have complex CPG networks for rapid adaptation in many types of situations. Engineers already know how to build *simple* CPG-based circuits that can enable robots to adapt their walking behavior over different types of terrain. These simple CPG circuits are usually discovered through several build-test cycles. We are proposing a design system for engineering *complex* CPG networks to provide a wide repertoire of behaviors to robots. This will enable building robots with a wider range of adaptive behaviors than is possible using finite-state machines and more rapid decision-making because there is no central processor and world-model to be consulted.

In the following we describe our preliminary work toward this goal.

## Biological Background and Objective

It is known that neurons behave like several types of electronic subsystems including free running oscillators, voltage controlled oscillators, frequency controlled oscillators and threshold logic gates (see Freeman, 2000; Silverston, 1985; Traub et al. 1999; Coombes and Bressloff, 2004). Central pattern generators (CPG) are neural oscillator circuits that are ubiquitous throughout the animal kingdom. They are used for controlling breathing, heart beating, walking, running, feeding, etc. They form to basis of the autonomic nervous system in many animals.

Animals as simple as jellyfish (canidarians) have a central pattern generator for controlling their swimming (Satterlie, 1985). In these simple animals the swimming is controlled by a ring network of coupled neurons located in the inner region of the animal. The network is made up of neurons that are about 22 microns in diameter. A 1:1 correspondence has been observed between the neuron spikes and epithelial synaptic potentials. Other, slightly more developed jellyfish (cubomedusae, Figure 1) have primitive eyes that directly control the swimming dynamics (Martin, 2002).

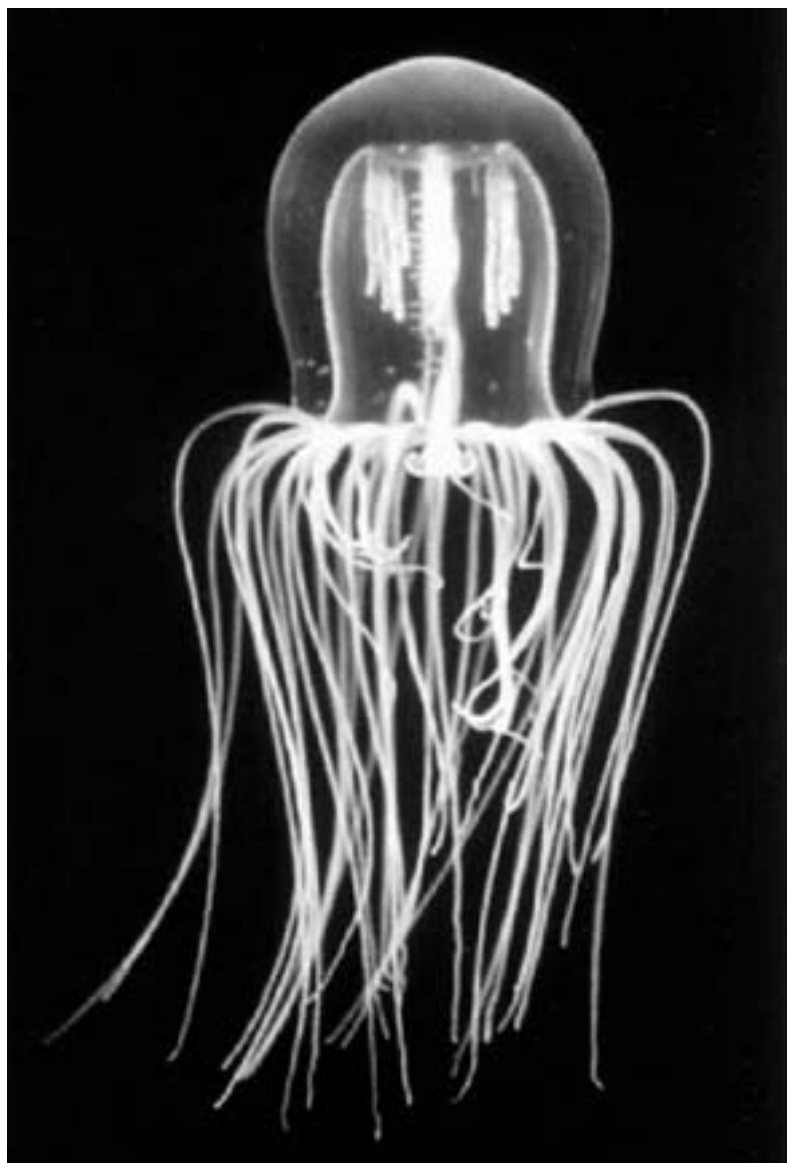

**Figure 1. Cubomedusae with photoreceptors just above the tentacles for directly controlling the dynamics of swimming and feeding (Photo from Martin, 2002).**

Examining more complex organisms, Ahn and Full (2002) have discovered that the cockroach extensively uses central pattern generators for walking and running. They removed the cockroach brain and found that the animal was able to navigate rough terrain because proprioceptors in the legs sent information directly into the CPGs that controlled the walking.

When organisms are evolving there are constraints on the design space. These constraints consist of common physical phenomena such as gravity, fluid dynamics, structural mechanics, etc. Once a “component” such as a central pattern generator has evolved it is utilized for as many applications as possible this includes building brains and autonomic nervous systems.

We hypothesize that computational systems can be built from CPGs - essentially new kinds of neural networks from rings of quasiperiodic oscillators. As with the jellyfish and the cockroach we are advocating the construction of neural hardware that does not use feedback learning but adapts rapidly to changing environment by linking sensors and actuators via CPG circuits. We want to know:

- What are the engineering limits of this as a computational paradigm?
- What do we have to do to exploit this technology to build robots with greater adaptability?

**Oscillators and Central Pattern Generators**

When a salamander walking through the grass encounters a beetle, it runs forward to capture the insect. The act of perceiving the insect, the act of changing from walking to running and the act of mouth opening are all done by very simple computation that does not require complex world models. The computation that these tasks comprise can come about through simple stimulus-response circuits of the autonomic nervous system. The adaptation comes about through real-time changes in the pulse patterns in central pattern generating neural circuits (Freeman, 2000; Silverston, 1985; Traub et al. 1999; Coombes and Bressloff, 2004). These circuits are a special class of ring oscillator.

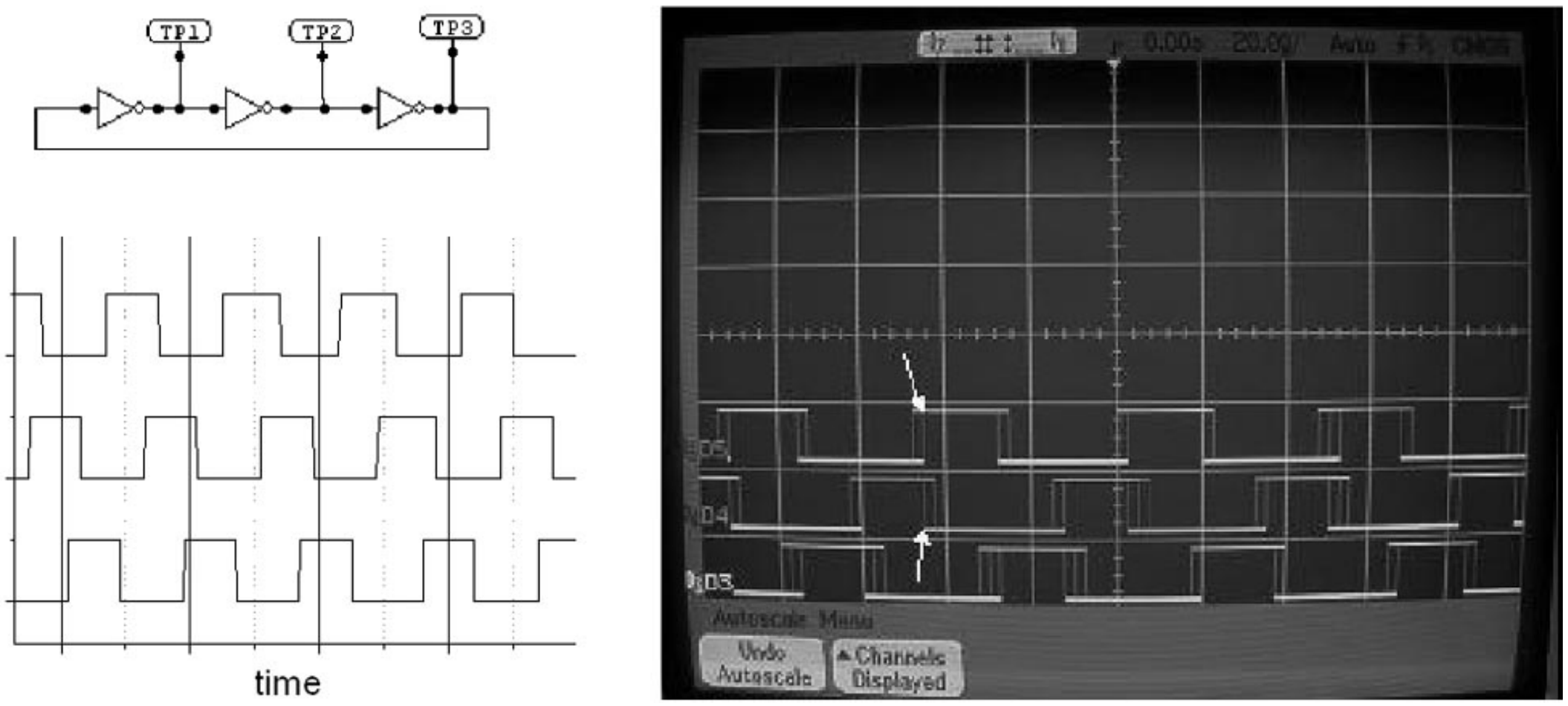

**Figure 2. Example of a ring oscillator. The output states alternate in time. The arrows point to "glitches" that result in phase noise.**

A typical ring oscillator can be constructed form a chain of an odd number of inverters that have been wired into a ring. Figure 2 shows an example. The input for node *n* is the output from node *n*-1. Since the logic gate, NOT, inverts the input it will produce a series of pulses that alternate in time. An even number of inverters in the circuit will not oscillate. The output state will be frozen at zero or one. Also notice in the figure that the pulse streams are not exactly aligned in time. They are not alternating precisely. When one gate is logic-high, its neighbor may also be high for a fraction of time. Also notice on the photo for the oscilloscope screen that there are glitches at the start and end of some of the pulses. Logic glitches, as they are called, are typically suppressed with Schmitt triggers. This suggests that if we built our oscillators with Schmitt triggers we would get more precise oscillators. If we put a time-delay element between the inverters we can get oscillations with an even number of nodes in the ring. An odd number of nodes will also oscillate but give more complicated behavior due to phase differences and subtle differences from component to component in the ring. (This topic will be discussed in detail later.)

Central pattern generators (CPGs) in biological organisms are a special type of oscillator used to control many autonomic functions. Figure 3 shows an example of an eight element CPG circuit to control the walking of a salamander gait. Synchronized pulses in the CPG circuit activate muscles that drive the legs. The CPG effectively controls the gait of the salamander. Humans and other complex organisms not only have brains, but also a complex autonomic nervous system. The main goal of our research is to develop an algebra of design – a design tool – for construction of the autonomic nervous system of robots.

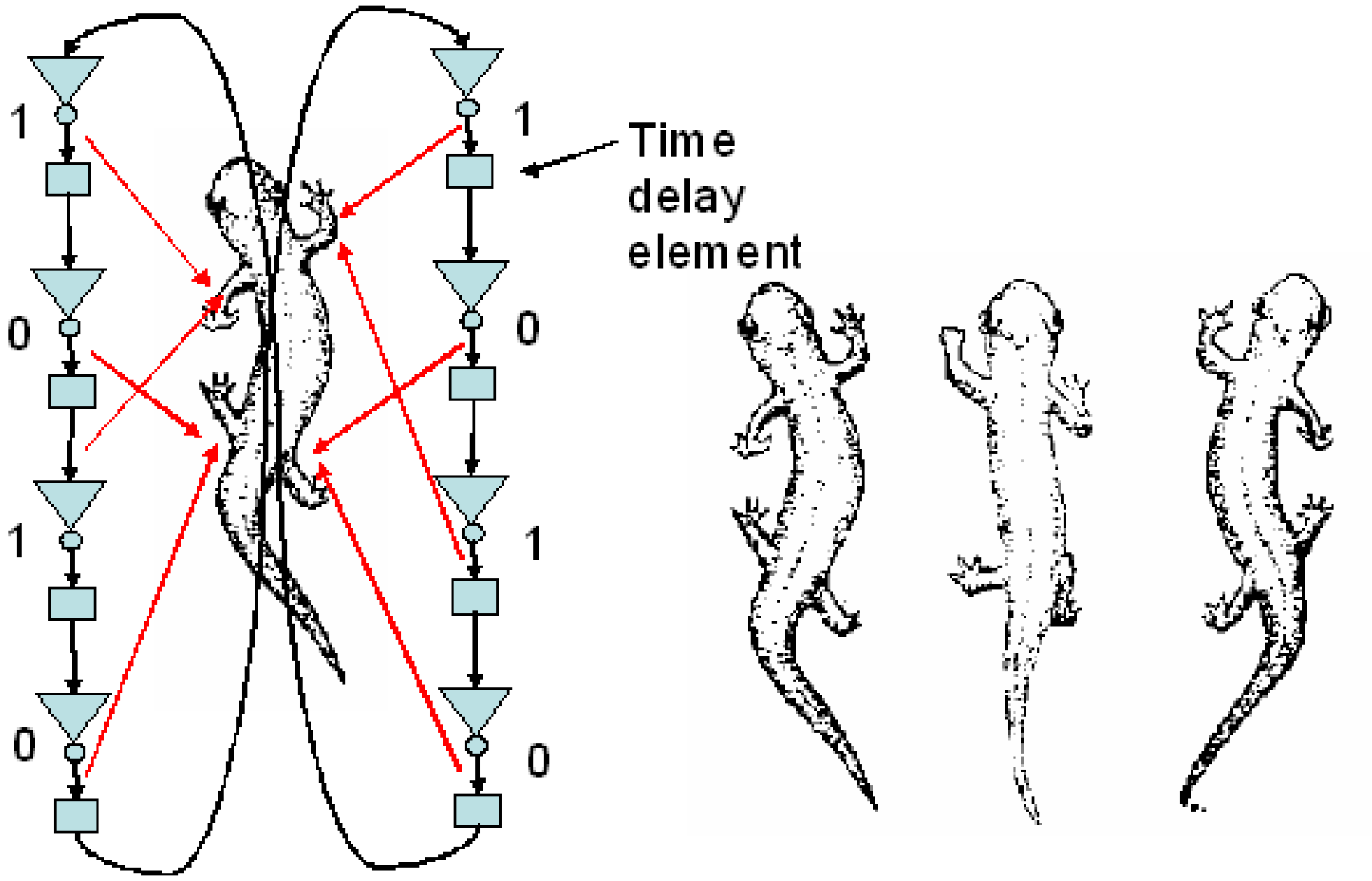

**Figure 3. Example of a central pattern generator circuit controlling the gait of a salamander.**

### Brief Literature Background

Some excellent papers have appeared in the literature discussing theoretical and experimental work on using central pattern generators for controlling walking robots. Hasslacher and Tilden (1995) and Still (2000) have described simple quadruped and hexapod robotic systems controlled by CPG circuits, not unlike those described by Ahn and Full (2000) in their work on cockroaches. Canavier et al. (1997, 1999) and Collins and Richmond (1994) describe CPGs modeled from van der Pol oscillators (Rietman, 1989), and Hoppensteadt and Izhikevich (1999, 2000) and Nishikawa et al. (2004) describe oscillatory associative memory.

Except for the work of Hoppensteadt and his colleagues, most of the technical papers have focused on CPGs for walking and controlling very simple robots. The modeling that has been done using van der Pol oscillators (Canavier et al., 1997, 1999; Collins et al., 1994) do not exhibit either the dynamics observed in simple ring oscillators constructed from hardware nor even the sigmoid behavior exhibited in biological neuronal circuits.

The van der Pol approach overlooks phase, frequency, and above all amplitude. The work of Hoppensteadt and his colleagues abstract away almost all details of CPGs and focus only on phase differences in coupled oscillators (Kuramoto, 1984).

The approach taken by hardware investigators is significantly different. Hasslacher and Tilden (1995) and Still (2000) use the ideas of building simple CPGs from ring oscillators with time delay between the nodes. Some very elegant machines have been built using a build-test approach. Figure 4, is a circuit diagram and photo of a robot built at Los Alamos National Labs. It is completely controlled with CPG circuits. Sensors send signals directly into the nodes that perform a summing-threshold operation. A signal can be injected directly into the CPG ring to change the robot gait. Another example is the Bio-Bug toys from Wow Wee Toy Company. The Bio-Bugs have simple recognition capability where they recognize similar and different species. The similar species will automatically flock and the different species will automatically fight each other. Mark Tilden claims this is emergent behavior and there are no digital CPU-based onboard circuits.

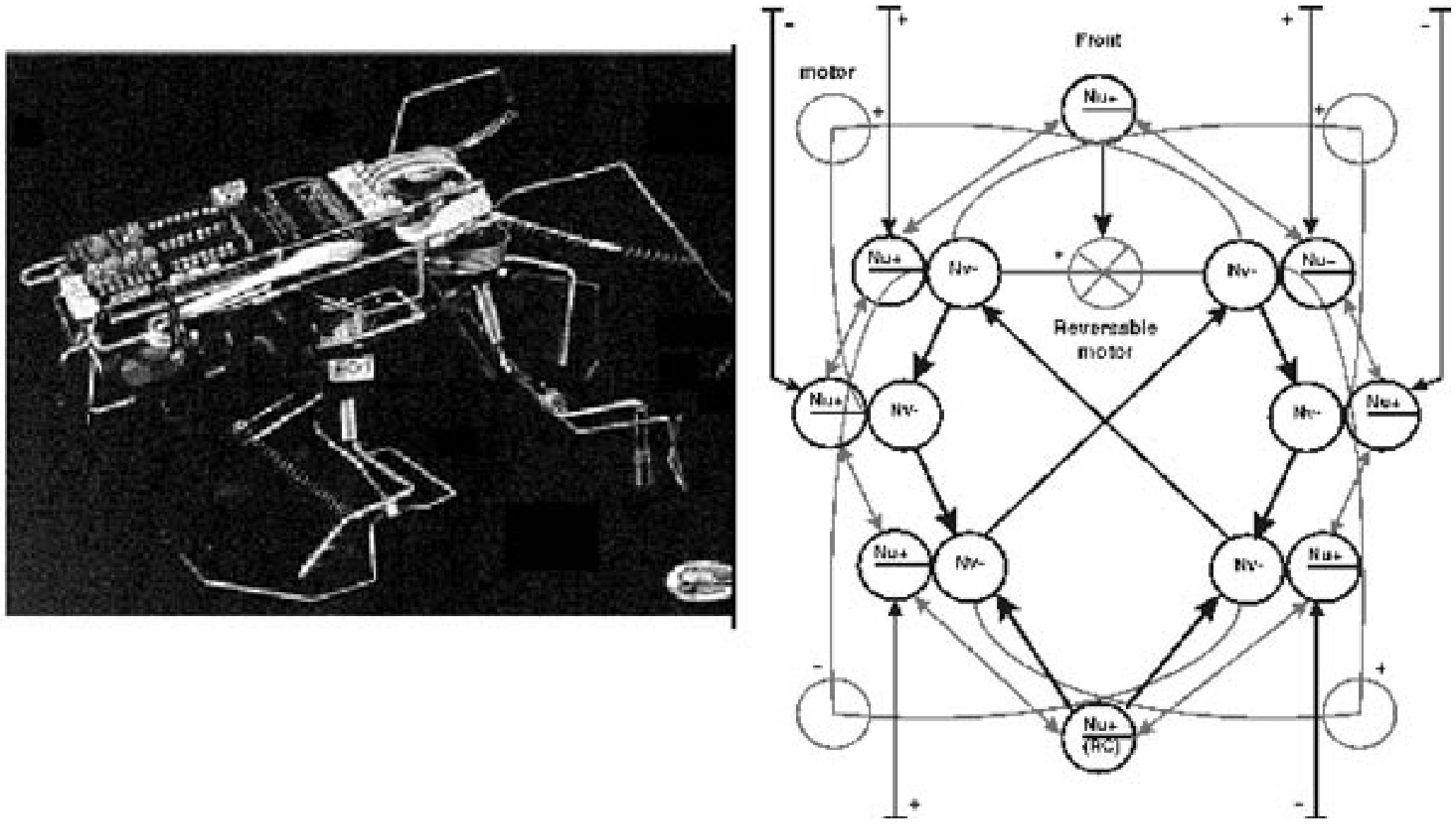

**Figure 4. CPG based robot built at Los Alamos National Laboratory.**

Emergent behavior and emergent computation are sometimes used as excuses to avoid deep analysis. We do not argue against the concept of exploiting emergent behavior, nor do we argue against using evolutionary algorithms for design of highly complicated systems for emergent behavior (Huelsbergen et al. 1999), but we do feel that a deeper understanding of engineering design using CPGs would be highly valuable in exploiting the adaptive capabilities of all types of robots (e.g. land, sea, air). Further, design-engineered CPG subsystems would be excellent starting points for evolutionary algorithms for evolving associative memories and more advanced computational engines.

The questions our research is attempting to answer are:

1) What are the fundamental building blocks that can be constructed from increasingly larger and larger CPG ring circuits?
2) What are the fundamental building blocks that can be assembled from these CPG ring circuits?
3) What are the reproducible dynamics for these circuits?
4) What are the design limitations to engineering these circuits?
5) What alternatives exist to circumvent the engineering limits?

**Theoretical Background – Simulations and Hardware**

In this theoretical background section we will focus only on what we consider the most relevant papers. We review some of the fundamental theoretical underpinnings that relate closely to hardware. We will not review the van der Pol oscillator or the Kuramoto oscillator. Though we take inspiration from biology we are not interested in biological realism in our models. Our interests are in hardware and developing new computational engines for robotics.

The basic neuron for our experiments is shown in Figure 5. It consists of a differentiating inverting Schmitt trigger. The capacitor is 0.01 microfarad and the resistor is 5.6 meaga-ohms. These values give a timing of 178 msec. (In the following we will simply reefer to this as - *the time constant*.) Granted this is a slow circuit, but that is exactly the type of speeds needed for controlling real world motors and actuators. The Schmitt trigger has an inherent voltage hysteresis. This provides some short-term memory to the component. The threshold voltage for the Schmitt trigger to fire depends on the actual component. We are using a CMOS 4093 (5 V power supply) with a hysteresis voltage typically of 0.7 V. When the input voltage is positive-going the device will trigger at 2.9 V. When the input voltage is negative-going the device will trigger at 2.2 V.

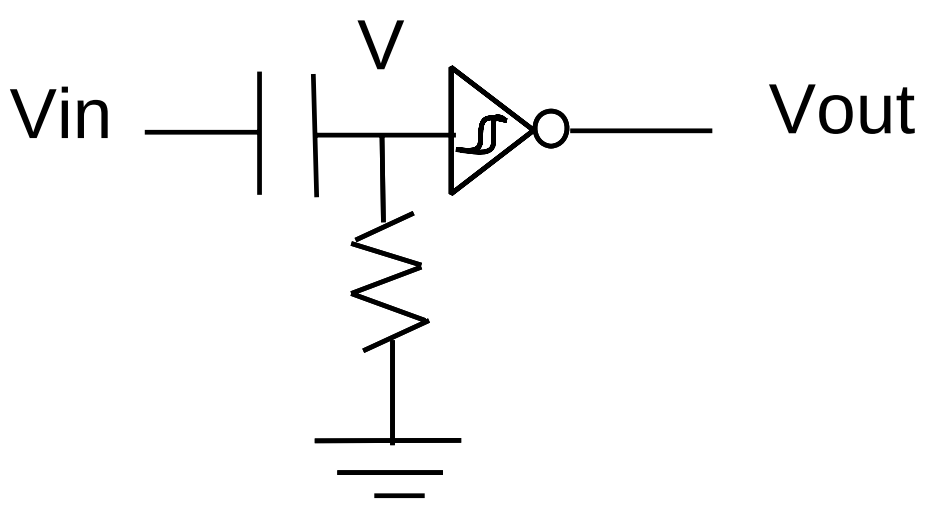

**Figure 5. Basic hardware neuron used in our experiments.**

In the relaxed powered-up state, in the absence of input, the output for the neuron shown in Figure 5 will be high. When there is an input sufficient to charge the capacitor the voltage at V will go high, the output will go low and the capacitor will discharge through the resistor to ground. When the neuron is low it is firing. After firing, via the discharge, the output will go high. The neuron will again enter the resting state. The input to the neuron can be expresses as $V_{in} = V_h[-u(t-t_0) + u(t-t_1)]$ where $V_h$ is the high output voltage (i.e. 5 V) and *u(t)* is a step function:

$$u(t) = \begin{cases} 0 & t < 0 \\ 1 & t \geq 0 \end{cases}$$

The behavior at node V is more complicated and is reviewed in Rietman et al. (2003). The firing time for the neuron is given by

$$t_2 - t_1 = -RC \ln\left[\frac{V_{thl}}{V_h\left(1 - \exp(-(t_1 - t_0)/RC\right)}\right]$$

This equation says that the firing time is dependent on the low voltage threshold, the magnitude of the input and the RC time constant. A long duration input pulse will cause a long output pulse and a short duration input pulse will cause a short duration output pulse. The effects are that short duration pulses will propagate faster through a network of these elements than long duration pulses.

If we now wire up rings of these neurons we find that even or odd numbers of them will oscillate. Typical ring oscillators must have an odd number of inverters to oscillate. In the case of the differentiating Schmitt trigger neurons a ring comprised of an odd number of neurons is wildly unstable (in hardware, not in simulation) due primarily to phase noise. In the case of ring circuits with an even number of these neurons, the circuit will oscillate but will exhibit a power law increase in the number of oscillatory states. It is found that these states are stable until disturbed whereupon they will shift to a different oscillatory state (see Rietman et al. 2003).

Since short duration pulses will propagate around the ring faster than long duration pulses, it is found in practice (i.e. hardware) that the short duration pulse can effectively catch-up to a slower propagating pulse. When this occurs they cause a frustrated situation (analogous to spin glasses in solid state physics) and the circuit must adapt to an entirely different oscillatory pattern. In large rings of >40 neurons these effects can actually be seen (provided each neuron is supplied with an LED output).

This behavior can be exploited (as is done for the Bio-Bugs ®). When a sensor sends a pulse (short or long duration) directly into the ring circuit controlling the walking gait the oscillatory pattern is disturbed and the circuit must settle to a different limit cycle. During this settling time the walking robot will appear to actually be adapting to a new gait. The settling time will typically be on the order of $N\tau$, where *N* is the number of nodes in the ring and $\tau$ is the average time constant for the neurons.

All logic gates, like that in Figure 5, can be thought of as high-gain amplifiers. We can model the neuron with the following sigmoid function:

$$\sigma(x) = 1/(1 + \exp(-\beta x)), \quad \beta > 0 .$$

In this equation the input to the neuron is given by *x*, and $\beta$ is the gain (amplifier gain). Often this equation is modified as follows for, so called, balanced sigmoids:

$$\sigma(x) = 2.0(1/(1 + \exp(-\beta x)) - 0.5)$$

Graphically this relation is shown in Figure 6. The neuron is able to accept positive or negative (excitatory or inhibitory) input. As the gain increases, the sigmoid function approximates a step function. With CMOS gates the sigmoid is very sharp. For the CMOS circuits we are using the hysteresis voltage is *about* 0.7 V. This uncertainty will manifest itself as Gaussian noise about the gain of the gate, $\beta + \varepsilon_\beta$, where $\varepsilon_\beta$ is the Gaussian noise.

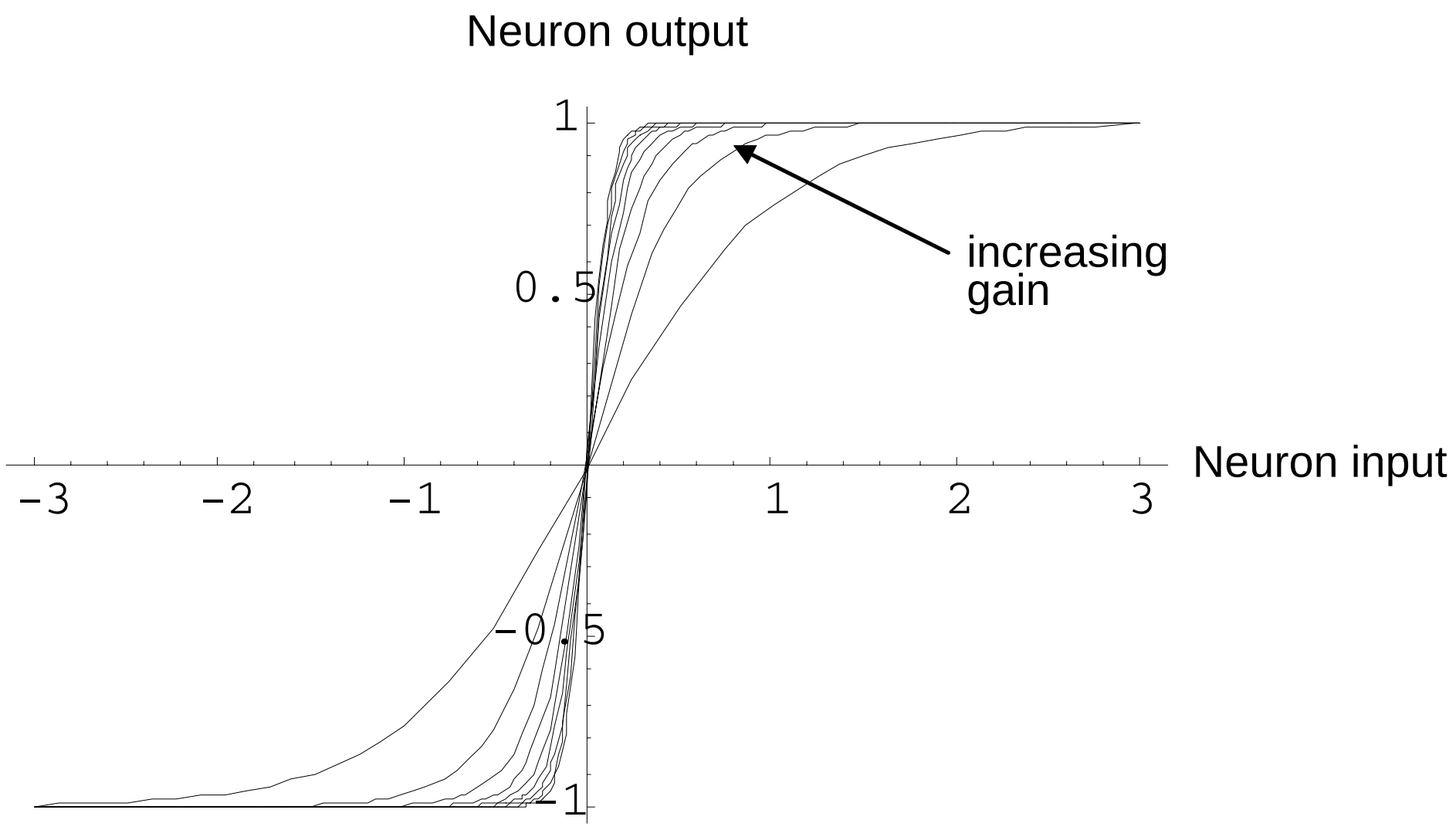

**Figure 6. Graphical display of the balanced sigmoid function at different levels of gain.**

A more detailed version of the neuron transfer function includes a bias term and its associated Gaussian noise, $\theta + \varepsilon_\theta$

$$\sigma(x) = 2.0(1/(1 + \exp(-(\beta x + \theta))) - 0.5)$$

With more than one input feeding into the neuron the inputs are simply summed as the product of the incoming signal and the synaptic weight from the first neuron to the neuron under consideration. The equation now becomes:

$$\sigma(x)_j = 2.0(1/(1+\exp(-(\beta_i \sum w_{ij} x_i + \theta_i))) - 0.5) \quad [1]$$

The equation above, without the noise terms, is the form used by most investigators modeling sigmoid based neural networks (Baum and Wang, 1992; Chapeau-Blondeau and Chauvet, 1992; Beiu et al, 1999; Amit, 1989). Though there are more recent papers with more complicated neuron architectures (e.g. Wilson, 1999), in most cases the dynamics can be approximated with this function (Eq. [1]).

We have selected a small set of literature sources for making a particular point about simulations, and because these sources describe the same type of networks we are interested in – central pattern generators.

The paper by Chapeau-Blondeau and Chauvet (1992) discusses stable oscillatory states in small ring networks consisting of only two or three neurons. In one example, two neurons are wired as shown in Figure 7a. In the second example three neurons are wired as shown in Figure 7b.

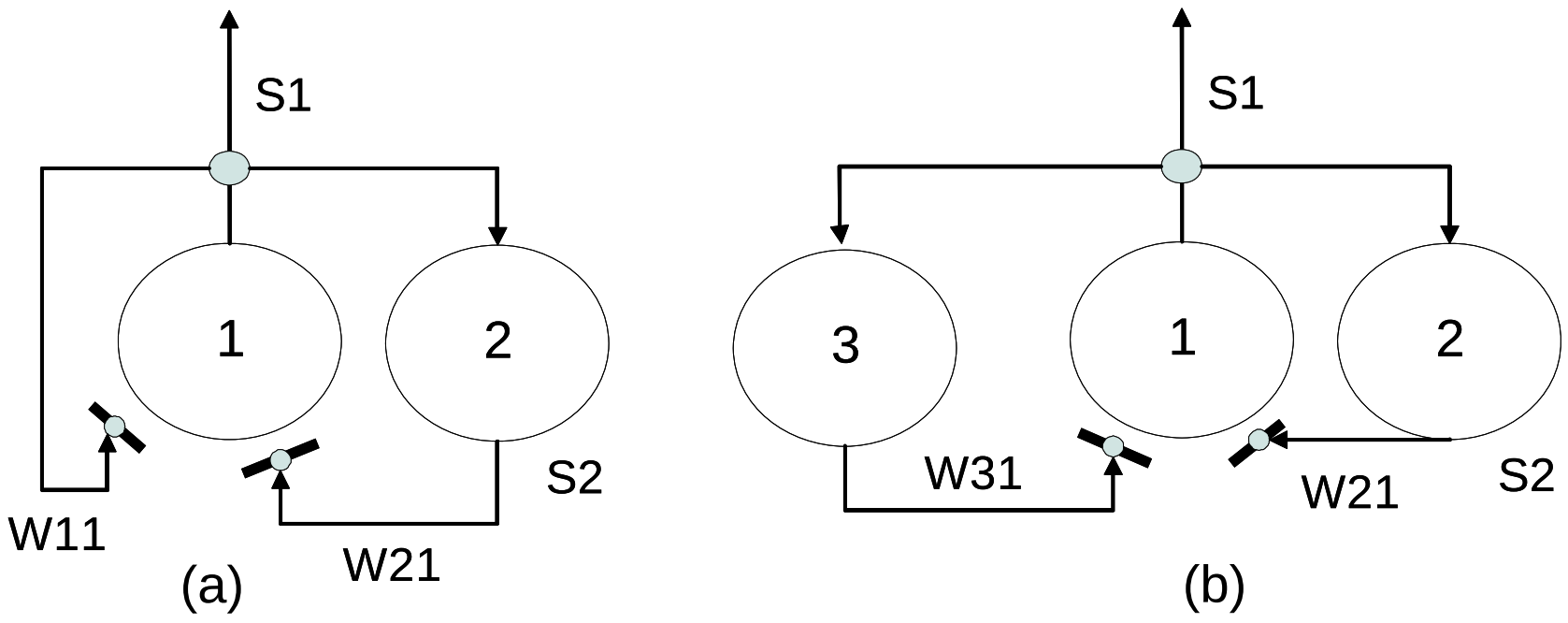

**Figure 7. Two example neural network described by Chapeau-Blondeau and Chauvet (1992). The direct arrows represent excitatory connections. The arrows on bars represent inhibitory connections.**

We verified the exact dynamics observed by Chapeau-Blondeau and Chauvet (1992). Figure 8 shows the simulation results for the following boundary conditions for the circuit shown in Figure 7b:

$$\beta_1 = \beta_2 = 7.0, \quad \beta_3 = 13.0, \quad \theta_1 = 0.5, \quad \theta_2 = 0.3, \quad \theta_3 = 0.7, \quad w_{12} = 1.0, \quad w_{31} = -0.8$$

As clearly evident, the circuit exhibits chaos. We argue, based on similar simulations, and understanding of the relations between these simulations (based on Eq. [1]) and hardware, that this is a special case. Before we present our arguments we will describe two additional technical papers.

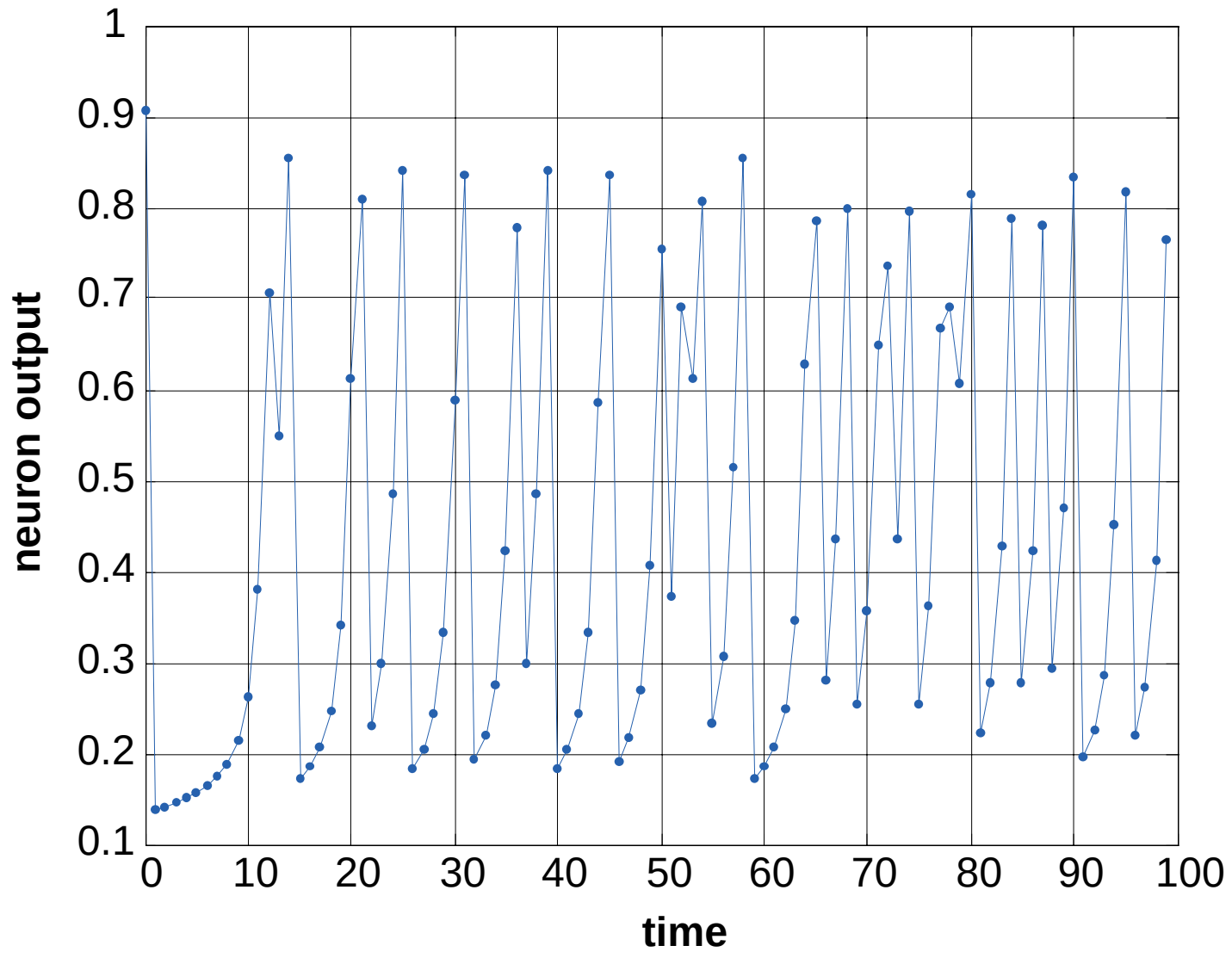

**Figure 8. Simulation results reproducing Chapeau-Blondeau and Chauvet (1992) for the circuit shown in Figure 7b.**

The papers by Blum and Wang (1992) and Beiu et al (1999) discuss eigenvalue analysis of ring networks built from the neuron described by Eq. [1]. They argue that when the weight matrix is given as:

$$W = \begin{bmatrix} 0 & 0 & 0 & \cdots & 0 & w_{n,1} \\ w_{1,2} & 0 & 0 & \cdots & 0 & 0 \\ 0 & w_{2,3} & 0 & \cdots & 0 & 0 & \cdots \\ \vdots & \vdots & \vdots & \cdots & \vdots & \vdots \\ 0 & 0 & 0 & 0 & w_{n-1,n} & 0 \end{bmatrix} \quad [2]$$

there will be stable limit cycles (not chaos). This matrix, of course, represents a nonsymmetric ring circuit. The elements are set to -1 (inhibitory connections). When the gain of the neurons, $\beta \leq 2$ (a soft sigmoid) will cause the circuit to settle to (0,0,0,…) where all neurons are in a quiescent state. At larger values of the gain (more realistic with respect to CMOS gates), a ring network of $n$ neurons will exhibit, by combinatorial considerations the periodic states:

$$\sum_{n=0}^{p} 2^n \binom{p}{n} = 3^p$$

and $3^n - 2^n - 1$ saddle points or metastable limit cycles.

Not all of these states need to be considered because of cyclic permutation for ring circuits. This series (see Appendix, Table 1) follows a 2-ary necklace function.

$$N(n,2) = \frac{1}{n}\sum_{i=1}^{v(n)} \phi(d_i)[F(d_i - 1) + F(d_i + 1)] \ , \qquad [3]$$

where $d_i$ are the divisors of *n* with $d_1 = 1, \quad d_{v(n)} = n$; $v(n)$ is the number of divisors of *n*; $\phi(n)$ is the totient function, F(.) is the Fibonacci sequence (see mathworld.wolfram.com for details). The totient function is also called the Euler Totient function is given as the number of positive integers less than *n*, which are relatively prime to *n* (Rietman, et al. 2003).

Though this infinite series (Eq. [3]) fits the data very well, the first 43 terms in the series also fits an exponential function

$$s = 0.066 \exp(n/2.188) + 1375.86$$

where *s* is the number of states and *n* is the number of nodes in the ring circuit. This is graphed in Figure 9. Also shown in the figure is a plot of the data on semilog scale. The main point of this is that the number of stable memory states increases exponentially as opposed to conventional networks (Hopfield networks) where the number of memory states is equal to 0.14*n*. Though this may seem like an unfair comparison because the Hopfield networks are constructed with the Hebb learning rule (Hopfield, 1982, 1984) and thus forced to minimize overlap between memory states. The Hebb rule has also been exploited by Amit (1988) and by Kleinfeld (1986) to build sequential state generators (multi-state oscillators) that do not have as many memory states as those we have constructed from Eqs. (1) and (2)).

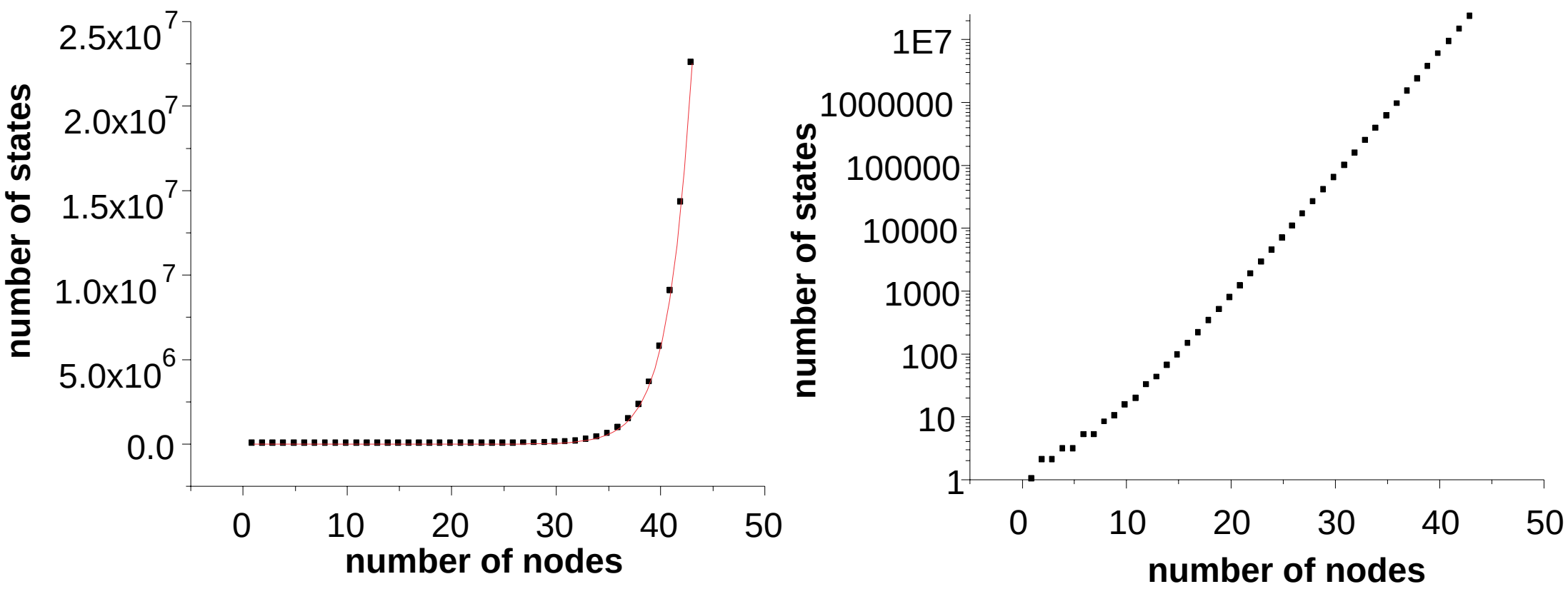

**Figure 9 Number of states for various sizes of rings according to the 2-ary necklace function Eq. [3].**

Hardware networks with an odd numbers of nodes will exhibit chaos. Typically for an odd number of inverters we could build a ring oscillator that, except for phase noise due to subtle transistor variations in the gates, would give stable oscillations. When building unidirectional ring circuits comprised of odd numbers of nodes of differentiating Schmitt triggers (Figure 5), the intrinsic hysteresis and RC component variation combined with the phase noise will not let the ring settle to a limit cycle – chaos will result. As shown in Figure 10, in simulations using Eq. [1] and [2] for a seven-node ring ($\varepsilon_\beta = 0, \quad \varepsilon_\theta = 0$), the output state is a quasiperiodic or multi-period limit cycle (13-cycle) – not chaos. The output displayed in the graph is the decimal equivalent of the zero-one state for the neurons (e.g. 120 = (1111000)). Unlike the hardware it is not chaotic, but rather quasiperiodic.

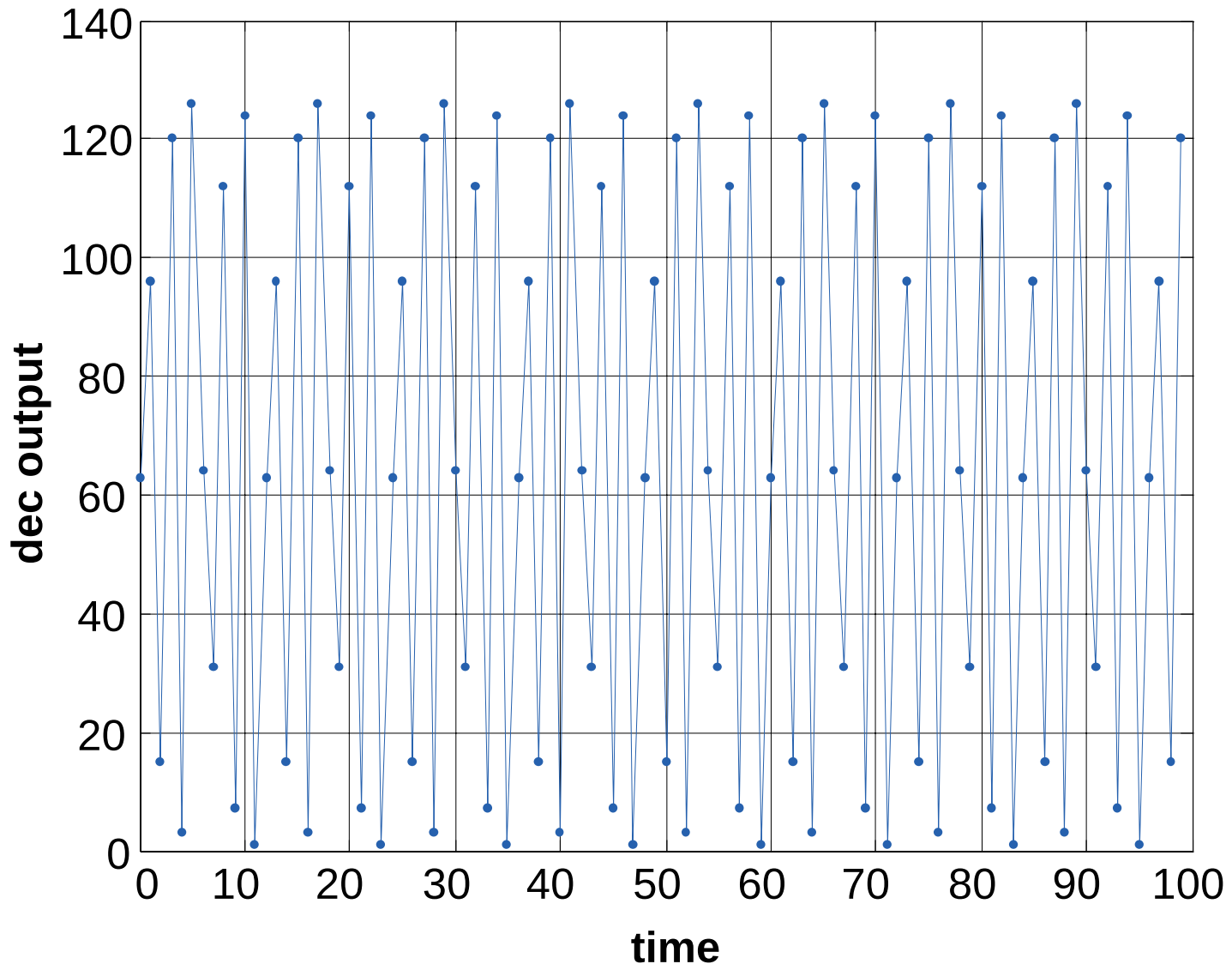

**Figure 10. Multi period limit cycle produced from a 7-node unidirectional ring built from Eq. [1] and [2].**

It is interesting that a 13-cycle is emitted by this 7-node ring network. Unless there was an inherent memory effect it should not be possible to produce a 13-cycle with a 7-node ring. At the most we should expect a 7-cycle.

For further comparison we examined 6-node and 8-node ring networks. As seen in Figure 11, in each case the periodicity of the limit cycle is identical to the number of nodes in the ring. This is in agreement with the analytical results of Blum and Wang (1992). The six- and eight-cycles are typical of those expected for these size networks. In each case the network was initialized with (-1,-1,-1, …).

Though these results seem to be reasonable and certainly correspond to the analytical approach of computing the eigenvalues of the weight matrix, in reality, when compared with hardware they are not. That is exactly our point. Much of the literature on these circuits does not correspond to real hardware circuits. Only when noise is added to the bias and gain do the simulations more closely match to hardware. Figure 12 shows simulation results for 6- and 8-node rings when the bias noise was set to 0 mean and 0.5 standard deviation and the gain noise set to mean 15 and standard deviation 3. The amount of noise is directly related to how quickly the network converges to either a stable limit cycle or a quasiperiodic state.

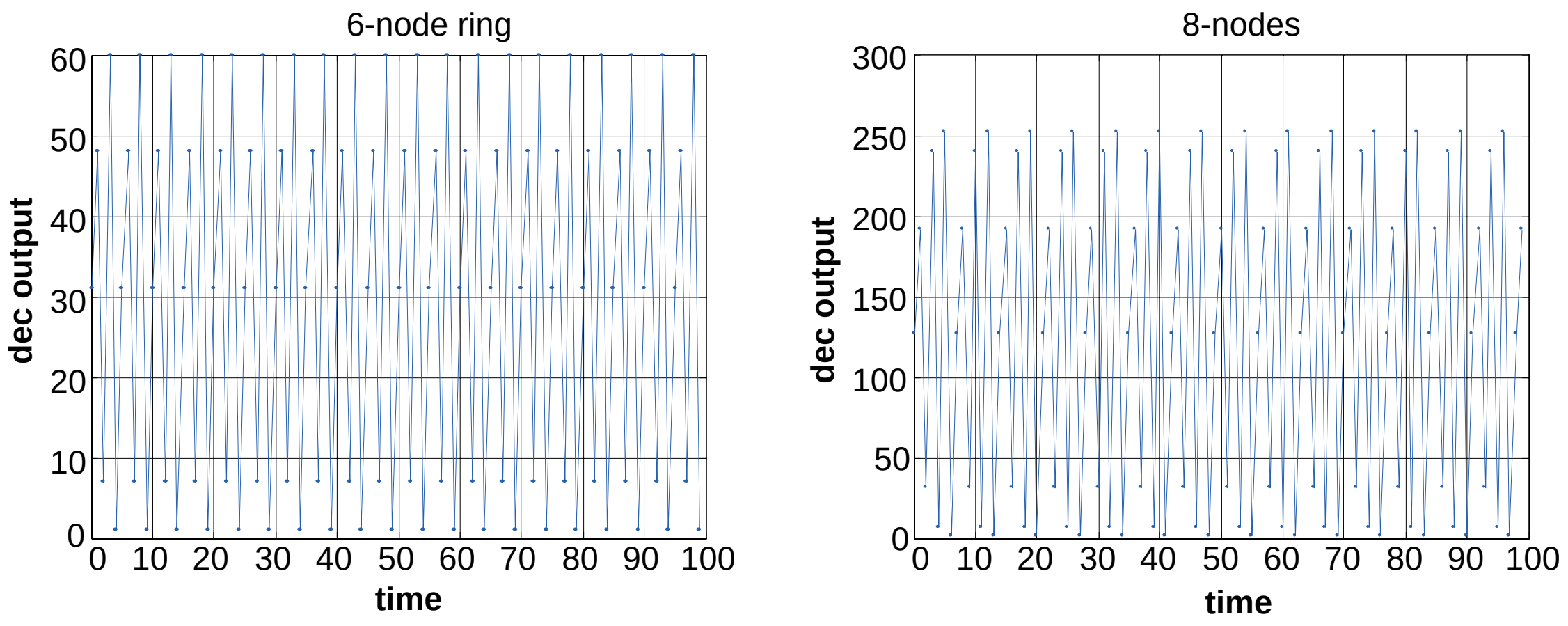

**Figure 11. Six-node and eight-node rings give rise to 6-cycle and 8-cycle respectively.**

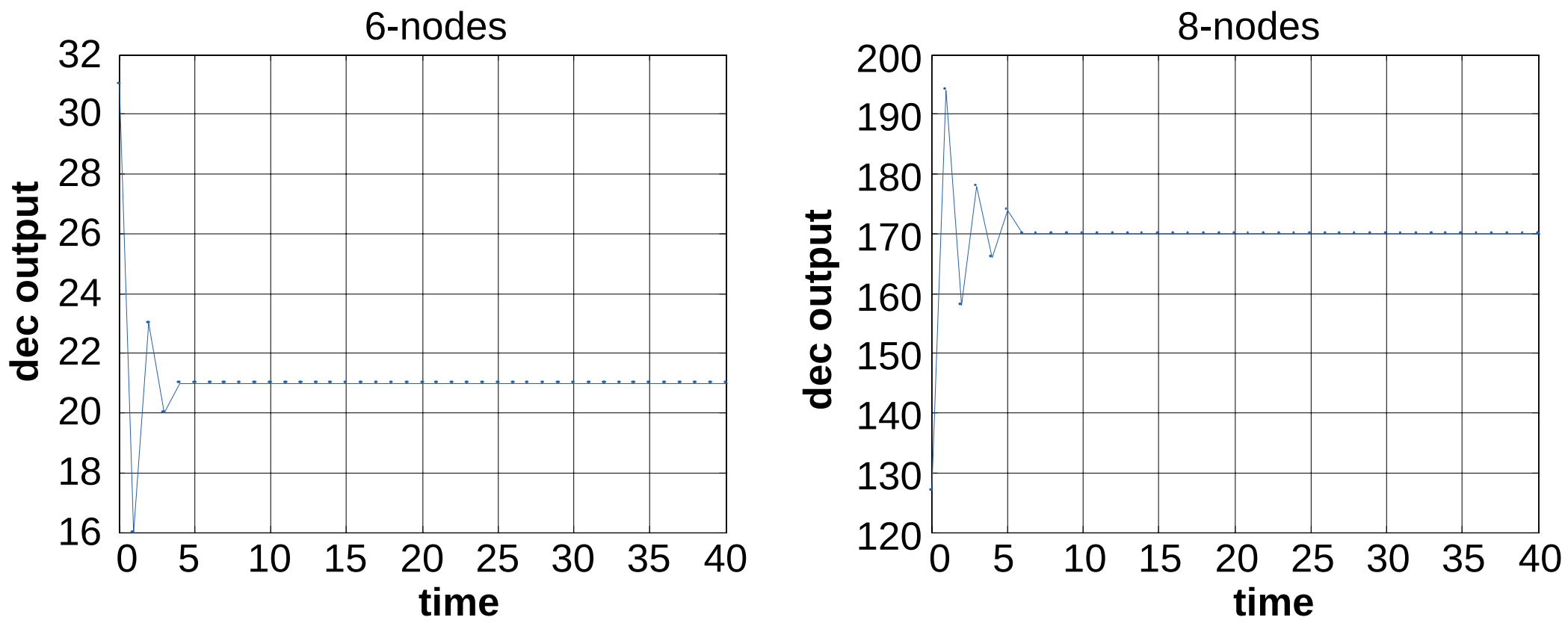

**Figure 12. Limit cycles (expressed as decimal equivalents) reached by 6- and 8-node rings when noise is added to the components in the circuit. Without noise the simulations do not approximate hardware.**

## Hardware Experiments

The basic hardware neuron used in our experiments is differentiating Schmitt trigger inverter (Figure 5). We are using CMOS 4093 NAND configured with the two inputs wired together. Forty eight individual neurons were wire wrapped on a Vector board to which additional components were added to allow monitoring the output state of each neuron with an LED, via a diode to reduce loading in the neural network. The output from the diode was amplified through a TTL inverting buffer (74LS04) and then through a current limiting resistor prior to the LED.

The neuron board also includes diode protected inputs controlled through a CMOS 4066 computer controlled analog switch. The outputs from the neurons are connected, via a ribbon cable, to a printed circuit board with latches also interfaced to a computer via a ribbon cable leading to a board with programmable interface chips (8255 PPI). A simplified block diagram is shown in Figure 13. A photo of the system is shown in Photo 1 in the Appendix and neuron board details are shown in Photo 2 in the Appendix.

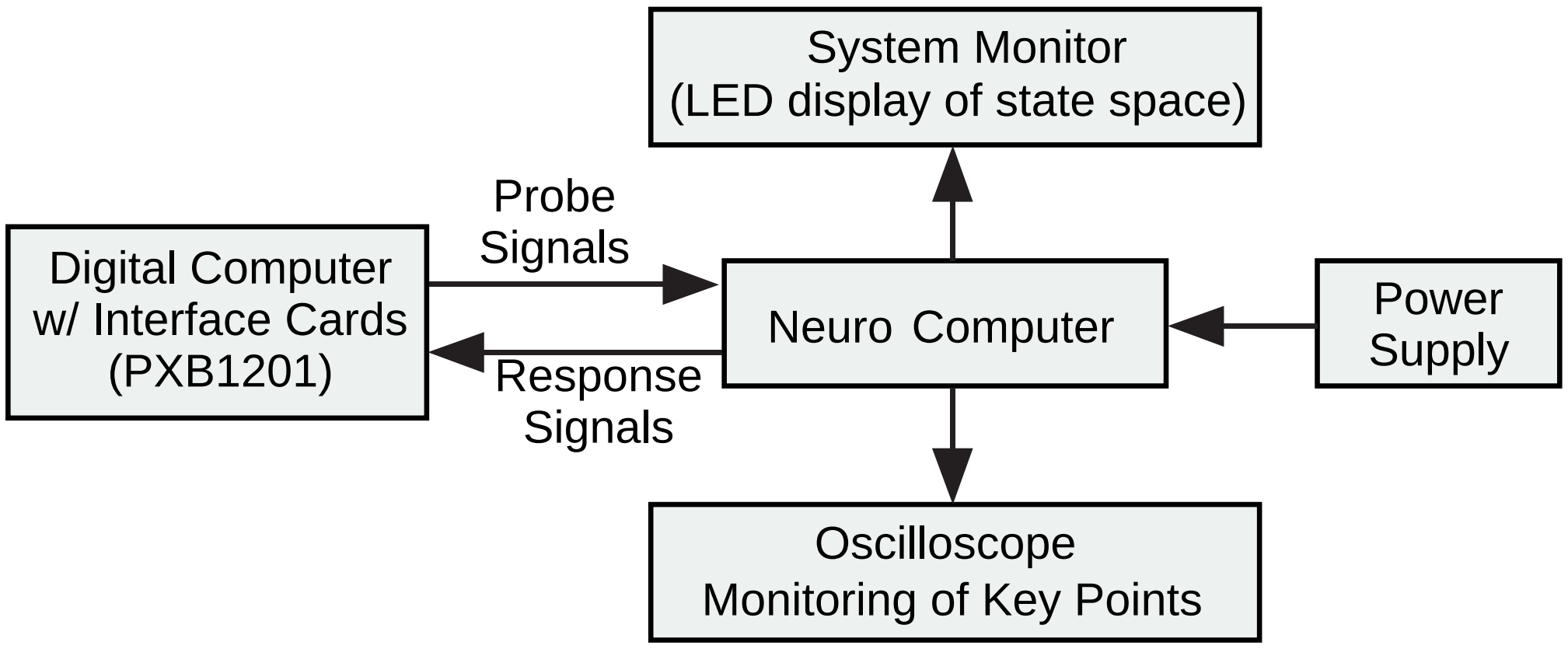

**Figure 13. Neurocomputer with computer control, LED display, and power supply.**

Our system was designed for maximum flexibility for configuration of neural circuits. As an example the resistors and capacitors in the differentiating configuration are in header sockets and can be swapped to reconfigure the neurons for integrate-fire type neurons. The massive numbers of cables leading to the proto-boards (see photos in Appendix) allows us to quickly configure ring circuits of varying sizes.

Figure 14 shows a schematic for an example of the ring circuits investigated in the project. Though the Figure shows a ring of 4-nodes we did not investigate that size ring. Instead, we investigated rings of 6-, 8-, 10-, 12-, 14-, and 16-nodes. The number of stable memory states in these networks is: 4, 7, 14, 30, 63 and 142 respectively. Table 1 shows the stable states for the 6- and 8-node rings. Because these are even number of nodes in the rings there is an additional stable state, a deep attractor – (00….000000). The memory capacity in these networks is far higher than the memory capacity of a conventional neural network, such as a Hopfield network. Our goal is to exploit this capacity.

Several important things should be noted about the bit strings shown in Table 1. First, there never are two 1s' next to each other. This is analogous to magnetic spins, they simply repel each other. Second, the bit string (000001), for example, is cyclically equivalent to (000010), (000100), etc. Note also that the bit string (010101) is equivalent to (101010). The number of zeros separating two ones is the primary observable. The maximum number of ones in the bit string is $n/2$, where $n$ (always an even number) is the number of nodes.

These observations suggest several things for the exploration of the phase space. First, it suggests we do not have to investigate the bit strings that are even number decimal equivalents. Second, it suggests that when the number of ones in the input bit string is $>n/2$ the ring is saturated. As an example, if we set the input state to (010111) for a 6-node ring the ring will quickly ($< n\tau$) settle to the limit cycle (010101).

We can summarize this and say that about one half of all input states will settle to the saturated state. This rule of thumb will prove useful later in explaining the observation of the basin boundaries obtained from the experimental measurements.

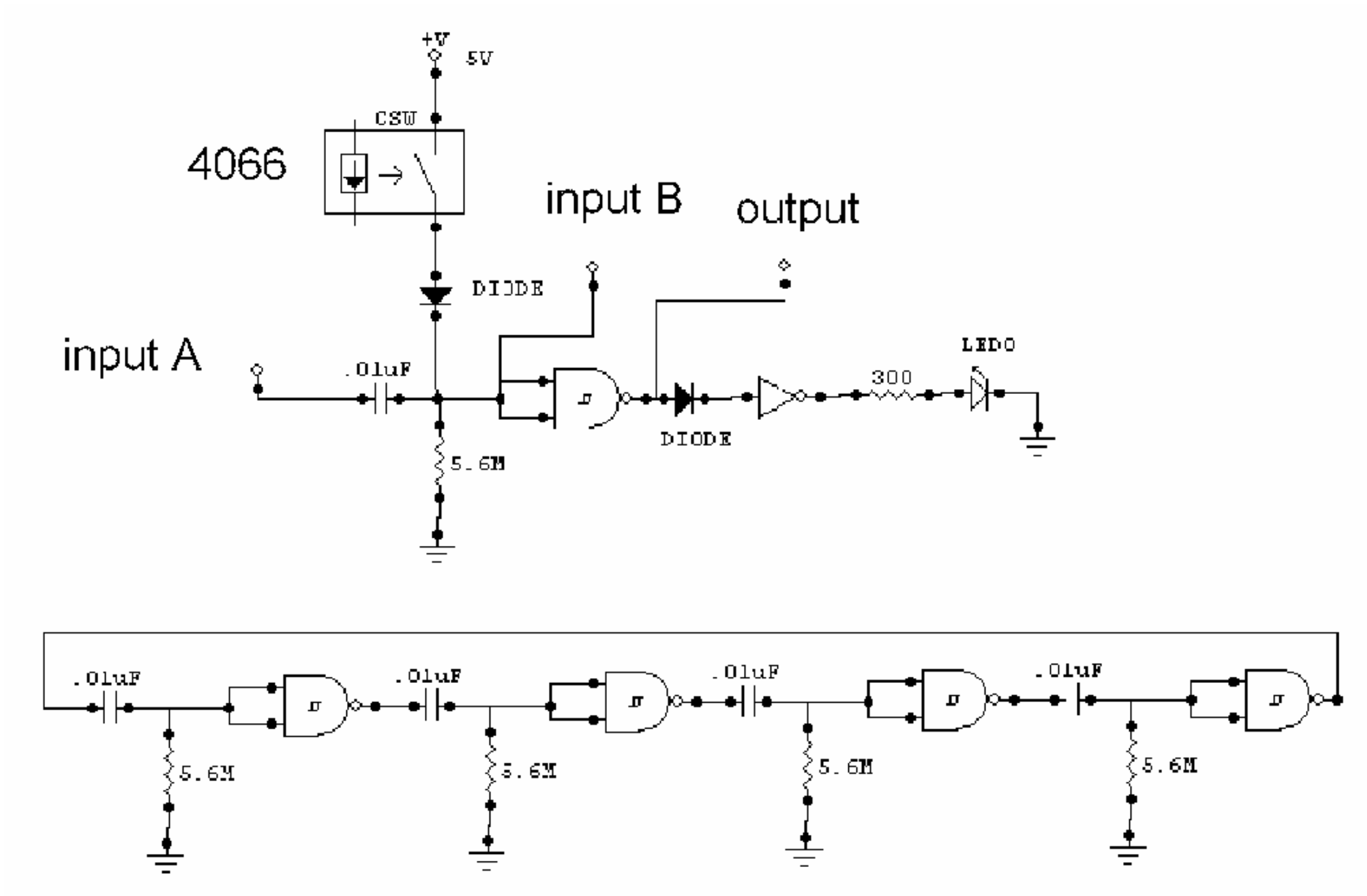

**Figure 14. Schematic of the individual neuron and a simplified schematic of a CPG circuit built from four neurons.**

| nodes | number of states | |
|---|---|---|
| | | 000001 |
| | | 000101 |
| | | 001001 |
| 6 | 4 | 010101 |
| | | 00000001 |
| | | 00000101 |
| | | 00001001 |
| | | 00010001 |
| | | 00010101 |
| | | 00100101 |
| 8 | 7 | 01010101 |

**Table 1. Memory states for the 6-node and 8-node ring circuits. The number of memory states is given by Eq. [3].**

Our experiments were conducted automatically and large sets of data were collected. A schematic for stimulating and monitoring the output state of each node in the ring is shown in Figure 15. The figure shows the programmable microswitches that were configured for the initial state of the neurons. All the switches were then opened simultaneously to allow the ring circuit to settle to the limit cycle. One-half second settling time was allowed prior to collecting 128 samples of all neuron states via the latches.

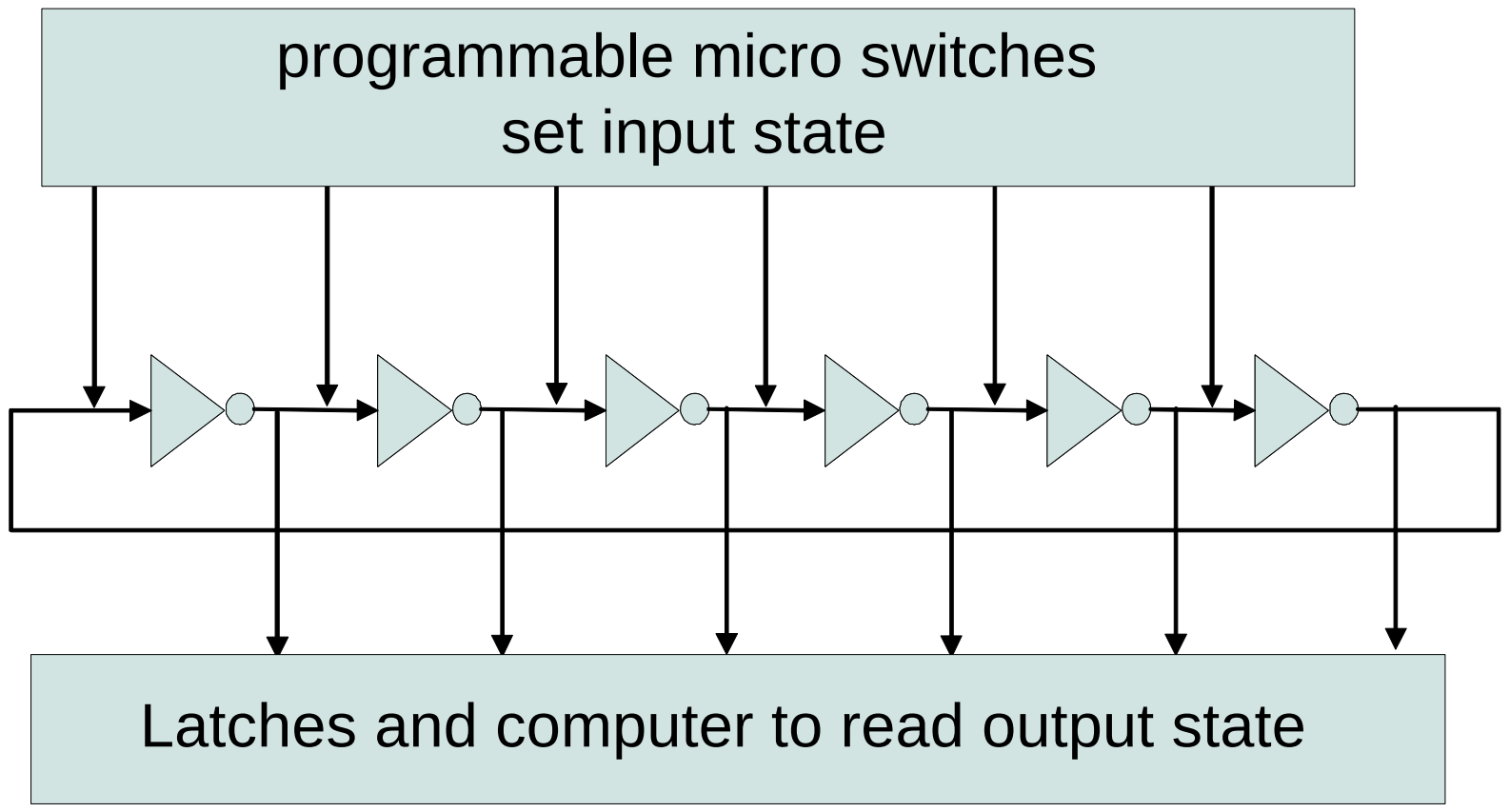

**Figure 15. Schematic of the typical experiment to determine the state space for the CPG circuits.**

**Ring Circuit Results and Preliminary Analysis**

The circuits, we constructed are dynamic systems; therefore, we describe the results in terms of fixed point attractors and limit cycles. When we set the initial configuration for a 6-node ring circuit with (000001) for example, we will have inadvertently selected a stable limit cycle for that architecture. As the circuit runs it will continue to display the cyclic permutation equivalent of (000001). Because of cyclic permutation of the outputs the limit cycle (000001) can be entered from several starting points. For example starting with (010000), decimal equivalent of 16, will be written as the lowest decimal equivalent 1. By simple cyclic permutation it is the same state.

If we start a 6-node machine in the state (000011) – decimal equivalent of 3 – the ring circuit will have to adapt. This is an unstable starting position because two 1s next to each other in a ring of inverters must settle to a different bit configuration. The two ones will effectively repel each other. It is important to realize that the circuits we have been experimenting with are *not* simple ring oscillators. Our circuits have an RC delay between the nodes. Consequently the node with the shortest RC delay will accelerate the pulse through that node and create a new bit string with the two ones separated by one zero (000101) – decimal equivalent 5. The end result looks as if the pulses repelled each other. This is a stable pattern. But if the resistors to ground (see Figure 5) are large (5-6 M ohm) there can easily be residual charge on the capacitors. Consequently the pattern (000101) will drift to (001001). If the resistor is on the order of 200 k ohm than the pattern (000101) is stable.

By the bit repelling argument the bit string (000011) is attracted to the limit cycle (000101) (state 5). All cyclic permutations of (000011) will be attracted to the same limit cycle. The set of bit strings attracted to state 5 is known as the basin boundary of that limit cycle. Similarly, other limit cycles, e.g. (010101), have their own basin boundary. Some of the limit cycles have larger basin boundaries than others. For example, for a six node machine and starting with any bit string with more than three ones the system will settle to the limit cycle (010101).

All of the attractors and basin boundaries for the six-node ring are shown in Figure 16. The figure on the left represents the decimal equivalent for the input and output and shows the number of samples that settled to the given output. The data represent six different, six-node machines, essentially an ensemble of results. Exhaustively evaluating the input/output dynamics there are a total of 64 possible input states (even numbers not included because of cyclic permutation equivalence). As seen in the graph the input state 1 settled to output state 1, thirty-two out of thirty-six times (considering all cyclic permutations of one bit). Of those four times that it did not settle to state 1 it settled to state 9 (001001) not 5 (000101). As pointed out above, in general, 9 is more stable than state 5 for two bits in the string. The basin boundaries for the other outputs are interpreted similarly. It should be noted, all these results are for 5.6 M ohm resistor to ground in the neuron circuit. This was a design decision to keep the time constant at values for real-world actuators.

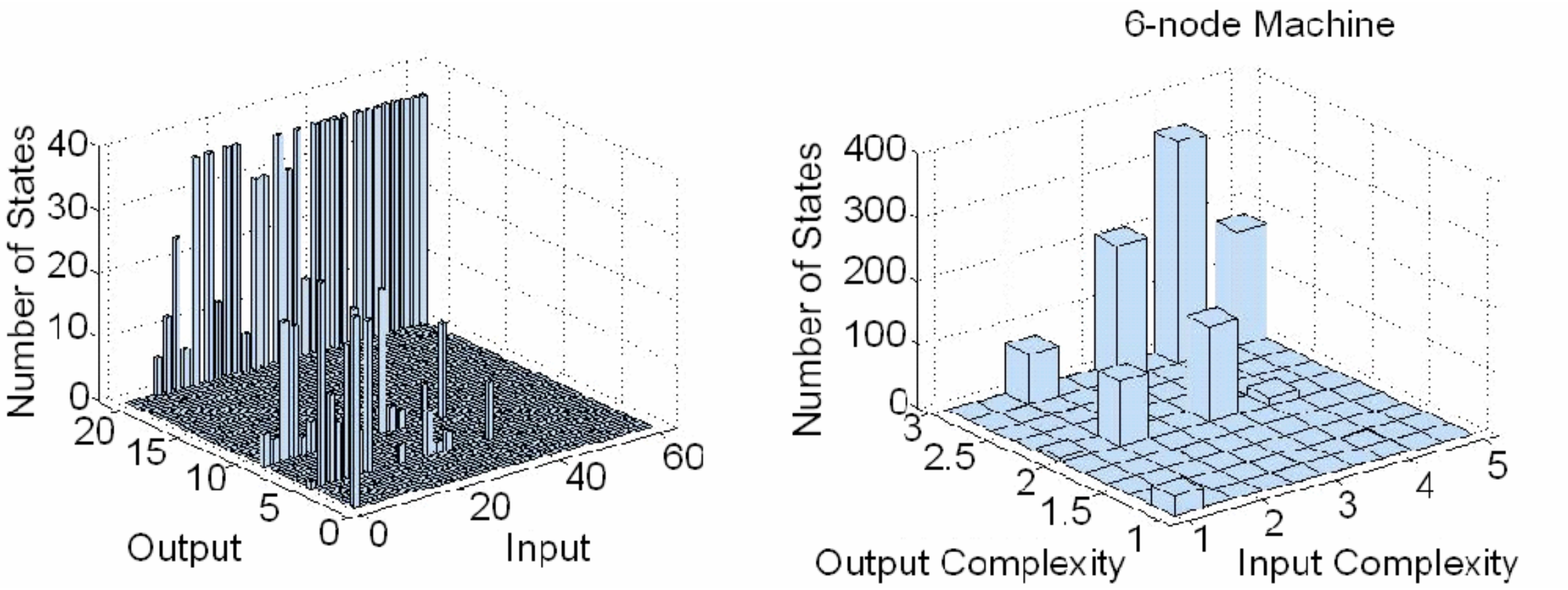

**Figure 16. Basin boundaries for 6-node machine. The numbers represent the lowest decimal equivalent states.**

The Figure on the right shows the same data represented as "complexity." We define "input complexity" is the number of ones in the initial bit string. The results show that most of the time when 3 or more 1s are on the input, the circuit will settle to (010101) – state 21. When two-ones are on the input string the machine usually settles to the limit cycle (000101) or (001001). Finally when there is only one, 1 in the input string the machine settles to the string (000001). Similar figures for 8-, 10-, 12-, 14-, and 16-node ring circuits are shown in the Appendix. They were not plotted on the same scale because each is a different ring circuit whose behavior we are presenting. They will be discussed in more detail from a group theory perspective in a later section.

The probabilistic appearance of these results is caused by residual charge on the capacitor. We are using large resistors to ground (5.6 M) to force the circuit dynamics to be slow enough to directly drive actuators and motors on robots. When these resistors are replaced with 200k we find that the circuits are deterministic and can be explained by group theory. The advantage of maintaining a group theory perspective is that the dynamics of the circuit can be described by deterministic algorithms, which will facilitate developing a design tool for exploiting these circuits for advanced computation tasks including building the autonomic nervous system of robots.

**Group Theory: Toward an Algebra of Design**

If we can cast the above results and further results in terms of group theory we will be able to use all of the conventional mathematical tools of abstract algebra to design systems. In order to use group theory to explain the observations we first need to either develop an algebra describing the groups observed or we need to find an similarity with known algebraic groups. In group theory this similarity is known as isomorphism. In a strict sense there has to be an exact one-to-one mapping between the elements of one group and the other group. This is called a bijection. In addition similar group operators should give rise to similar group elements within the allowed sets. In the following we will describe the groups for the 6-node and 8-node ring circuits. Later in this paper we describe the groups for rings up to 16-nodes.

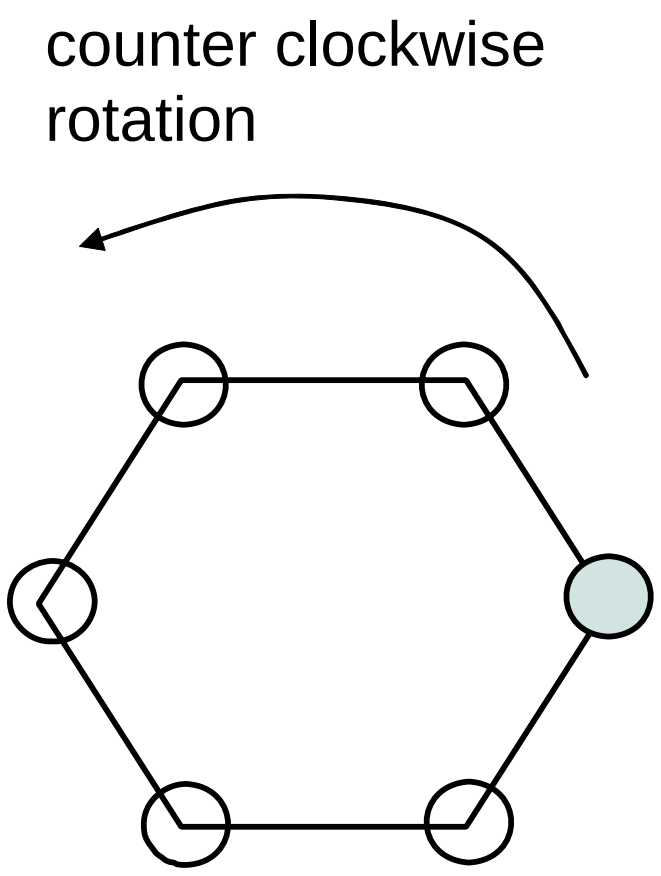

**Figure 17. Diagram of six-node CPG oscillator.**

Consider the 6-node ring with only one bit active (000001). This is shown in Figure 17 as a hexagon with one circle filled. If the active bit is traveling in the counter clockwise direction we can represent the transitioning bit string as follows:

$$(000001)\xrightarrow{r}(000010)\xrightarrow{r}(000100)\xrightarrow{r}(001000)\xrightarrow{r}$$
$$(010000)\xrightarrow{r}(100000)\xrightarrow{r}(000001)$$

After six-rotations, *r*, the ring is in the same configuration as when we started. (This is said to be a six-cycle in the terminology of dynamic systems.) Symbolically we can represent this as:

$$\underline{1}\xrightarrow{r}\underline{2}\xrightarrow{r}\underline{4}\xrightarrow{r}\underline{8}\xrightarrow{r}\underline{16}\xrightarrow{r}\underline{32}\xrightarrow{r}\underline{1}$$

Where the numbers represent the decimal equivalent of the bit string and they are underlined to remind us that these are group symbols not numbers to be manipulated as numbers. This string of elements interspersed with a rotation operation represents the elements for the group and the main operation. We represent this group by $G_1^6$ where the superscript reminds us that the group is for six-node rings and the subscript is the lowest decimal equivalent of the bit string in this group.

In addition to the rotation operator we have potential mirror-plane and mirror-axis symmetry operations. We will represent the set of all possible symmetry operators for a group as {m}. The $G_1^6$ group can now be written as $G_1^6$={1, 2, 4, 8, 16, 32, r, {m}}.

The group $G_1^6$ describes only one of the possible cyclic groups within the 6-node ring circuit. Since there are four stable oscillatory states in the 6-node machine there are four groups total. For the moment we will drop the symmetry operators and reintroduce them as needed. The full set of all the groups $g^6 = \{G_1^6, G_5^6, G_{91}^6, G_{21}^6\}$ is given as:

$$g^6 = \begin{cases} G_1^6 : \{\underline{1} \xrightarrow{r} \underline{2} \xrightarrow{r} \underline{4} \xrightarrow{r} \underline{8} \xrightarrow{r} \underline{16} \xrightarrow{r} \underline{32} \xrightarrow{r} \underline{1}\} \\ G_5^6 : \{\underline{5} \xrightarrow{r} \underline{10} \xrightarrow{r} \underline{20} \xrightarrow{r} \underline{40} \xrightarrow{r} \underline{33} \xrightarrow{r} \underline{17} \xrightarrow{r} \underline{5}\} \\ G_9^6 : \{\underline{9} \xrightarrow{r} \underline{18} \xrightarrow{r} \underline{36} \xrightarrow{r} \underline{9}\} \\ G_{21}^6 : \{\underline{21} \xrightarrow{r} \underline{42} \xrightarrow{r} \underline{21}\} \end{cases}$$

The above set of mappings shows cyclic permutations from rotation operations on the individual states $\underline{s}$ represented as decimal equivalent. The $G_1^6$ and $G_5^6$ groups are said to be of order 6. The $G_9^6$ group is fourth order and the group $G_{21}^6$ is second order. The similarities between group theory and conventional dynamics are obvious. The two 6 order groups are 6-cycles. The one third-order group is a three-cycle and the second-order group is a two-cycle.

The rotation operator (applied once) for each group is different

$$\{G_1^6, G_5^6, G_9^6, G_{21}^6\} = \left\{\frac{\pi}{3}, \frac{\pi}{3}, \frac{2\pi}{3}, \pi\right\}$$

As the number of rotations needed to return to the starting state decreases for a given group, the periodicity increases – e.g. a two-cycle is faster than a 6-cycle (a walking robot will walk faster with 2-cycle than 6-cycle). Similarly, as the number of rotations needed to return to the starting state decrease, the order of the group decreases and the symmetry increases. As we point out later, a symmetry phase transition occurs during sensor fusion and ring coupling.

Before discussing the isomorphism and writing the group tables for the $g^6$ group we will introduce the well known cyclic groups from abstract algebra.

The cyclic group $C_6$ consists of the decimal numbers {0, 1, 2, 3, 4, 5) and the operation

$$\rho((a+b)\,\mathrm{mod}(6))$$

where $\rho$ is the operator that adds two elements in the group *a*, *b* and then applies the modulus operation.

| C6 group table | | | | | | |
|---|---|---|---|---|---|---|
| **a\b** | **0** | **1** | **2** | **3** | **4** | **5** |
| **0** | 0 | 1 | 2 | 3 | 4 | 5 |
| **1** | 1 | 2 | 3 | 4 | 5 | 0 |
| **2** | 2 | 3 | 4 | 5 | 0 | 1 |
| **3** | 3 | 4 | 5 | 0 | 1 | 2 |
| **4** | 4 | 5 | 0 | 1 | 2 | 3 |
| **5** | 5 | 0 | 1 | 2 | 3 | 4 |

**Table 2. The $C_6$ group table.**

The $C_6$ group table is shown in Table 2. The first row in the table is the elements of the group. The first column is the elements of the group, written in the same order as the elements in the first row. The actual arrangements of the elements in the first row/column are not important. The first row is *a* the first column is the element *b*, for the operator $\rho$ . The elements in the table are generated by the operator, just like a multiplication table.

The index *p* of a cyclic group $C_p$ is given by

$$Z(C_p) = \frac{1}{p}\sum_{k|p}\varphi(k)a_k^{p/k}$$

where $k \mid p$ means *k* divides *p*; $\varphi(k)$ is the totient function discussed with Eq. [3] and **Z** is the set of integers (http://mathworld.wolfram.com/CyclicGroup.html ). Recall, Eq. [3] generates the number of stable states for each of the CPG circuits.

| G1 group table | | | | | | |
|---|---|---|---|---|---|---|
| | **<u>1</u>** | **<u>2</u>** | **<u>4</u>** | **<u>8</u>** | **<u>16</u>** | **<u>32</u>** |
| **<u>1</u>** | <u>1</u> | <u>2</u> | <u>4</u> | <u>8</u> | <u>16</u> | <u>32</u> |
| **<u>2</u>** | <u>2</u> | <u>4</u> | <u>8</u> | <u>16</u> | <u>32</u> | <u>1</u> |
| **<u>4</u>** | <u>4</u> | <u>8</u> | <u>16</u> | <u>32</u> | <u>1</u> | <u>2</u> |
| **<u>8</u>** | <u>8</u> | <u>16</u> | <u>32</u> | <u>1</u> | <u>2</u> | <u>4</u> |
| **<u>16</u>** | <u>16</u> | <u>32</u> | <u>1</u> | <u>2</u> | <u>4</u> | <u>8</u> |
| **<u>32</u>** | <u>32</u> | <u>1</u> | <u>2</u> | <u>4</u> | <u>8</u> | <u>16</u> |

**Table 3. The $G_1^6$ group table.**

The group table for the $G_1^6$ group is given in Table 3. Similar to the $C_6$ group table the elements are written across the first row and first column. Recall the underline is to remind us that these are symbols not numbers. We define the group operation $\otimes$ according to the following mapping:

$$
\begin{aligned}
(G_1^6, \otimes) &\leftrightarrow (\mathrm{Z}_6, \oplus) \\
\underline{1} &\leftrightarrow 0 \\
\underline{2} &\leftrightarrow 1 \\
\underline{4} &\leftrightarrow 2 \\
\underline{8} &\leftrightarrow 3 \\
\underline{16} &\leftrightarrow 4 \\
\underline{32} &\leftrightarrow 5
\end{aligned}
\quad .
$$

This maps the CPG group $G_1^6$ to the first nonnegative integers $\mathrm{Z}_6$ in the cyclic group $C_6$.

By the defined mapping we have established an isomorphism between these two groups

$$G_1^6 \cong C_6$$

The other isomorphisms that exist for the $g^6$ set of groups are

$$
\begin{aligned}
G_5^6 &\cong C_6 \\
G_9^6 &\cong C_3 \\
G_{21}^6 &\cong C_2
\end{aligned}
$$

When we reintroduce the mirror symmetry set $\{m\}$ we will find that the G groups are isomorphic with conventional dihedral groups.

In order to use these ideas with concepts such as sensor fusion and CPG ring coupling we need to define operators $\Phi_{nr}$ that transform one group into another group. We have selected the symbol $\Phi$ to represent this operator because it will remind us that phase is important. Let the subscript on the operator represent the number of rotations when the pulse is injected. Then we can write all of the allowed operations on the groups and their results. The system of equations describing group transformations for sensor fusion into the 6-node ring are shown in the Appendix.

Referring to the $g^6$ sensor fusion operator group transformations in the Appendix consider as an example $\Phi_2 : G_1^6 \to G_5^6$. This equation says that when the CPG circuit has one, 1 cycling through the ring and if a pulse of duration equal to the time constant of the neurons is injected at rotation 2 (subscript to operator) this will be the equivalent of initializing the ring circuit with (000101) or decimal 5. So the circuit is transformed to the $G_5^6$ group. Explicitly this would be written as (000100) + (000001) → (000101).

As another example consider $\Phi_{23} : G_1^6 \to G_9^6$. This relationship says that when a pulse of two time constants are injected at rotation positions 2 and 3 into a 6-node circuit with a signal already at position 0 (always the assumed initial state), the circuit pulse pattern will transform to $G_9^6$. Explicitly this would be written as (000001) + (000110) → (0001001). The other equations in the Appendix are interpreted similarly.

Counting the number of group transformations that settle to the respective limit cycles indicated in the equations we find there is only 1 transformation to $G_1^6$. There are 11 transformations to $G_5^6$; 9 transformations to $G_9^6$; and 64 transformations to $G_{21}^6$, with a total of 85 equations. The fractions respectively are 1%, 13%, 10%, and 75%. From our experimental work where we initialized the ring circuit to a particular state and let it settle to the limit cycle we obtain empirically the probability graph shown in Figure 18. The observed probabilities are very close to those computed by the group transformations. These results also explain the observed basin boundaries shown in Figure 16.

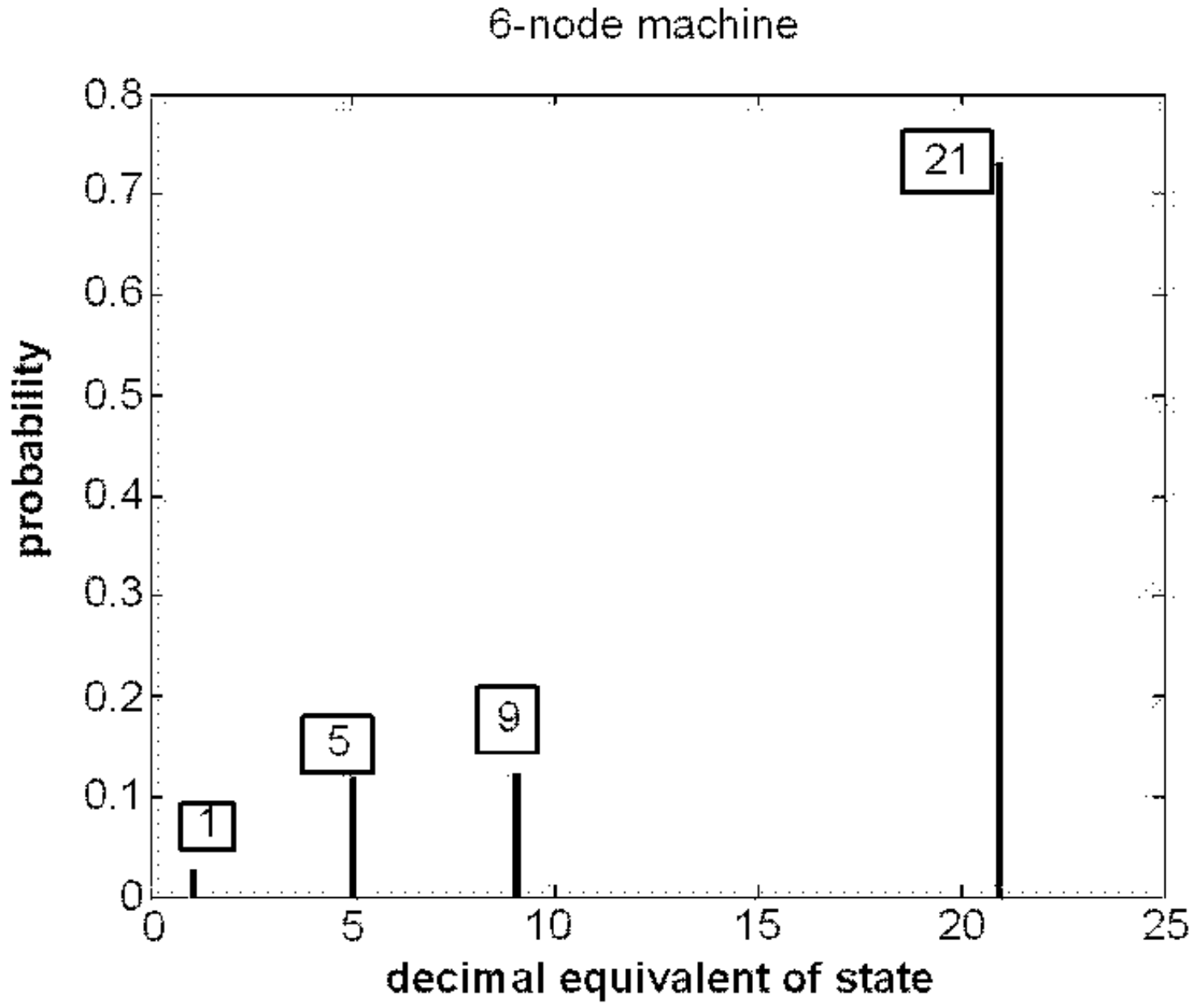

**Figure 18. Probability of settling to the indicated limit cycles in a 6-node CGP circuit.**

### 8-node Ring Group Theory

Recall our notation for the state of a machine is represented by the lowest decimal equivalent of the binary number, and to insure that we do not confuse this with decimal numbers we underline the state. The 8-node machine has 7 stable states other than zero.

These states are ($\underline{1}$, $\underline{5}$, $\underline{9}$, $\underline{17}$, $\underline{21}$, $\underline{37}$ and $\underline{85}$). The group operations for each of these states are given by the following series of rotation equations.

$$
\begin{aligned}
&G_1^8 : \underline{1} \xrightarrow{r} \underline{2} \xrightarrow{r} \underline{4} \xrightarrow{r} \underline{8} \xrightarrow{r} \underline{16} \xrightarrow{r} \underline{32} \xrightarrow{r} \underline{64} \xrightarrow{r} \underline{128} \\
&G_5^8 : \underline{5} \xrightarrow{r} \underline{10} \xrightarrow{r} \underline{20} \xrightarrow{r} \underline{40} \xrightarrow{r} \underline{80} \xrightarrow{r} \underline{160} \xrightarrow{r} \underline{65} \xrightarrow{r} \underline{130} \\
&G_9^8 : \underline{9} \xrightarrow{r} \underline{18} \xrightarrow{r} \underline{36} \xrightarrow{r} \underline{72} \xrightarrow{r} \underline{144} \xrightarrow{r} \underline{33} \xrightarrow{r} \underline{66} \xrightarrow{r} \underline{132} \\
&G_{17}^8 : \underline{17} \xrightarrow{r} \underline{34} \xrightarrow{r} \underline{68} \xrightarrow{r} \underline{136} \\
&G_{21}^8 : \underline{21} \xrightarrow{r} \underline{42} \xrightarrow{r} \underline{84} \xrightarrow{r} \underline{168} \xrightarrow{r} \underline{81} \xrightarrow{r} \underline{162} \xrightarrow{r} \underline{69} \xrightarrow{r} \underline{138} \\
&G_{37}^8 : \underline{37} \xrightarrow{r} \underline{74} \xrightarrow{r} \underline{148} \xrightarrow{r} \underline{41} \xrightarrow{r} \underline{82} \xrightarrow{r} \underline{164} \xrightarrow{r} \underline{73} \xrightarrow{r} \underline{146} \\
&G_{85}^8 : \underline{85} \xrightarrow{r} \underline{170}
\end{aligned}
\tag{4}
$$

The rotation operator (applied once) for each group is different:

$$\left\{G_1^8, G_5^8, G_9^8, G_{17}^8, G_{21}^8, G_{37}^8, G_{85}^8,\right\} = \left\{\frac{\pi}{4}, \frac{\pi}{4}, \frac{\pi}{4}, \frac{\pi}{2}, \frac{\pi}{4}, \frac{\pi}{4}, \pi\right\}$$

Group tables can be written and we can define the group operations $\otimes$ according to the following mapping:

$$
\begin{aligned}
(G_1^8, \otimes) &\leftrightarrow (Z_8, \oplus) \\
(G_5^8, \otimes) &\leftrightarrow (Z_8, \oplus) \\
(G_9^8, \otimes) &\leftrightarrow (Z_8, \oplus) \\
(G_{17}^8, \otimes) &\leftrightarrow (Z_4, \oplus) \\
(G_{21}^8, \otimes) &\leftrightarrow (Z_8, \oplus) \\
(G_{37}^8, \otimes) &\leftrightarrow (Z_8, \oplus) \\
(G_{82}^8, \otimes) &\leftrightarrow (Z_2, \oplus)
\end{aligned}
$$

where $Z_i$ represents the appropriate cyclic group with its standard operations (Barnard and Neill, 1996; Whitelaw, 1995).

The sensor fusion equations are in the Appendix. Group operations for rings of size 10, 12, 14, and 16 are given in the Appendix.

**Relating this Group Theory to Robot Walking Gait Control**

As an example of applying these group theory concepts to a simple robot consider the hexapod robot shown in Figure 19. The CPG contains nodes numbered N1 to N6. Each of the nodes is connected (via diodes) directly to stepper motors labeled M1 to M6. In addition nodes N1 and N6 receive sensor signals from S1 and S2 respectively. Each motor simply controls a rigid leg that is able to swing a few degrees about its shoulder when it receives a pulse.

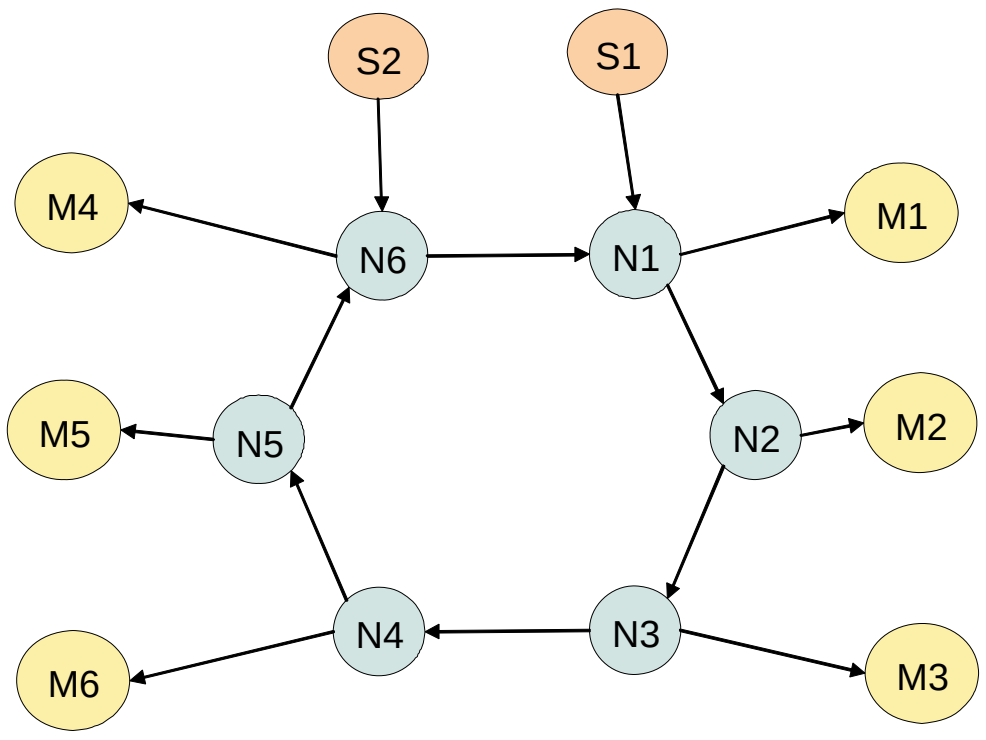

| pattern | (000001) | slow walkng | | | | |
|---|---|---|---|---|---|---|
| time | M1 | M2 | M3 | M4 | M5 | M6 |
| 1 | 1 | 0 | 0 | 0 | 0 | 0 |
| 2 | 0 | 1 | 0 | 0 | 0 | 0 |
| 3 | 0 | 0 | 1 | 0 | 0 | 0 |
| 4 | 0 | 0 | 0 | 1 | 0 | 0 |
| 5 | 0 | 0 | 0 | 0 | 1 | 0 |
| 6 | 0 | 0 | 0 | 0 | 0 | 1 |
| | | | | | | |
| pattern | (000101) | medium speed walking | | | | |
| time | M1 | M2 | M3 | M4 | M5 | M6 |
| 1 | 1 | 0 | 1 | 0 | 0 | 0 |
| 2 | 0 | 1 | 0 | 1 | 0 | 0 |
| 3 | 0 | 0 | 1 | 0 | 1 | 0 |
| 4 | 0 | 0 | 0 | 1 | 0 | 1 |
| 5 | 1 | 0 | 0 | 0 | 1 | 0 |
| 6 | 0 | 1 | 0 | 0 | 0 | 1 |
| | | | | | | |
| pattern | (001001) | medium speed walking | | | | |
| time | M1 | M2 | M3 | M4 | M5 | M6 |
| 1 | 1 | 0 | 0 | 1 | 0 | 0 |
| 2 | 0 | 1 | 0 | 0 | 1 | 0 |
| 3 | 0 | 0 | 1 | 0 | 0 | 1 |
| | | | | | | |
| | | | | | | |
| | | | | | | |
| | | | | | | |
| pattern | (010101) | fast walking | | | | |
| time | M1 | M2 | M3 | M4 | M5 | M6 |
| 1 | 1 | 0 | 1 | 0 | 1 | 0 |
| 2 | 0 | 1 | 0 | 1 | 0 | 1 |

**Figure 19. Simple hexapod robot architecture with six motors.**

If the robot is ambling along with walking gate $G_5^6$ when the phase is (001010) and receives a pulse on S1 of duration equal to the time constant of the neurons then the robot will turn slightly to the left at the next time constant and continue with walking gait $G_5^6$. If the robot now receives another signal of the same duration and from the same sensor then it will turn slightly to the left and change its walking gait to $G_9^6$. Not significantly different in speed. Another sensor signal of the same duration and from the same sensor will cause the robot to change its walking gait to $G_{21}^6$, a much faster pace.

Though this example is rather simple, it does show that we can predict behavioral changes in the robot as a result of signals. As we will show in the final report, similar analysis can be done with coupled CPG circuits and thereby enable design of behaviors from first principles.

## Conclusions

We have described the full dynamics of rings of inhibitory neurons from an experimental and theoretical perspective. We have found that when the hardware circuits have large (e.g. 5 Mohm) value resistors, needed for controlling real-world actuators, the residual charge will force the circuit to behave in a probabilistic manor. When the resistors are smaller (e.g. 200 kohm) the probabilistic nature of the circuits disappears and the full dynamics can be explained with group theory.

Thus we have outlined that certain ring circuits of neurons can act as reproducible building blocks for central pattern generators and that these CPG circuits can have a very large number of stable oscillatory states. We have described the engineering limits of these circuits as the RC time constant.

Future work with these circuits will consist of networks constructed from the rings and an elaboration of the group theory to include mirror symmetry.

**Acknowledgments**

This material is based upon work supported by the U.S. Army and DARPA under Contract No. W311P4Q-06-C-0079. Any opinions, findings and conclusions or recommendations expressed in this material are those of the author(s) and do not reflect the views of the U.S. Army or DARPA.

**References**

Ahn and Full (2002), “A Motor and Brake: Two Leg Extensor Muscles Acting at the Same Joint Manage Energy Differently in Running Insect,” *Journal of Experimental Biology*, 205, 379-389.

Amit (1988), “Neural Networks Counting Chimes,” *Proc. Natl. Acad. Sci. USA*, 85, 2141-2145.

Amit, D. J. (1989), *Modeling Brain Function, The World of Attractor Networks*, Cambridge U. Press.

Arken, (1998), *Behavior-Based Robotics*, MIT Press, Cambridge.

Barnard, T. and Neill, H. (1996), *Mathematical Groups*, Teach Yourself Publishing.

Baum and Wang (1992), “Stability of Fixed Points and Periodic Orbits and Bifurcations in Analog Neural Networks,” *Neural Networks*, 5, 577-587.

Beiu et al (1999), “On the Reliability of the Nervous Nets,” Los Alamos National Laboratory, LA-UR-98-3461.

Bekey (2005), *Autonomous Robots*, MIT Press, Cambridge.

Canavier et al. (1997), Phase Response Characteristic of Model Neurons Determine Which Patterns are Expressed in a Ring Circuit Model of Gait Generation,” *Biol. Cybern*. 77, 367-380.

Canavier et al. (1999), "Control of Multistability in Ring Circuits of Oscillators," *Biol. Cybern*, 80, 87-102.

Chapeau-Blondeau and Chauvet (1992), "Stable, Oscillatory, and Chaotic Regimes in the Dynamics of Small Neural Networks with Delay," *Neural Networks*, 5, 735-743.

Collins and Richmond (1994), "Hard-wired Central Pattern Generators for Quadrupedal Locomotion," *Biol. Cybern*, 71, 375-385.

Coombes and Bressloff (editors) (2004), *Bursting and the Genesis of Rhythms in the Nervous System*, World Scientific.

Freeman (2000), *Neurodynamics, An Exploration in Mesoscopic Brain Dynamics*, Springer, New York.

Gill, J. (1977), "Computational Complexity of Probabilistic Turing Machines," *SIAM J. Computing*, 6(4), 675-695.

Hasslacher and Tilden (1995), "Living Machines," *Robotics and Autonomous Systems*, 15, 143-169.

Hopfield (1982), "Neural Networks and Physical Systems with Emergent Collective Computational Abilities," *Proc. Natl. Acad. Sci. USA*, 79, 2554-2558.

Hopfield (1984), "Neurons with Graded Response have Collective Computational Properties like those of two-state Neurons," *Proc. Natl. Acad. Sci. USA*, 81, 3088-3092.

Hoppensteadt and Izhikevich (1999), "Oscillatory Neurocomputers with Dynamic Connectivity," *Physical Rev. Lett.* 82, 2983-2986.

Hoppensteadt and Izhikevich (2000), "Pattern Recognition via Synchronization in Phase-Locked Loop Neural Networks," *IEEE Trans. on Neural Networks*, 11, 734-738.

Huelsbergen et al. (1999), "Evolution of Astable Multivibrators *in Silico*," *Evolvable Systems: From Biology to Hardware*, Sipper, Mange and Perez-Uribe, editors, 66-77, Springer, New York, NY.

Kleinfeld, (1986) "Sequential State Generation by Model Neural Networks," *Proc. Natl. Acad. Sci. USA*, 83, 9469-9473.

Kuramoto (1984), *Chemical Oscillations, Waves, and Turbulence*, Springer-Verlag, New York.

Martin (2002), "Photoreceptors of Cnidarians," *Can. J. Zool.*, 80, 1703-1722.

Murray, A. and Tarassenko, L. (1994), *Analogue Neural VLSI: A Pulse Stream Approach*, Chapman & Hall, London, 1994.

Nishikawa, et al. (2004), "Capacity of Oscillatory Associative-Memory Networks with Error-Free Retrieval," *Physical Rev. Lett.*, 92, 108101.

Park, S. (2005), *A Programming Language for Probabilistic Computation*, Ph.D. Dissertation, Carnegie Mellon University, CMU-CS-05-137.

Rietman (1988), *Experiments in Artificial Neural Networks*, WindCrest Tab, Blue Ridge Summit, PA.

Rietman (1989), *Exploring Parallel Processing*, WindCrest Tab, Blue Ridge Summit, PA.

Rietman, et al. (2003), "Analog Computation With Rings of Quasiperiodic Oscillators: The Microdynamics of Cognition in Living Machines," *Robotics and Autonomous Systems*, 45, 249-263.

Satterlie (1985), "Central Generation of Swimming Activity in the Hydrozoan Jellyfish *Aequorea Aequorea*," *J. of Neurobiology*, 16, 41-55.

Silverson (1985), *Model Neural Networks and Behavior*, Plenum, New York.

Still (2000), *Walking Gait Control for Four-Legged Robots*, PhD dissertation, Swiss Federal Institute of Technology, Zurich.

Traub et al. (1999), *Fast Oscillations in Coritical Systems*, MIT Press, Cambridge.

Wilson, H. R. (1999), *Spikes Decisions and Actions, Dynamical Foundations for Neuroscience*, Oxford U. Press.

Whitelaw, T. A. (1995), *Introduction to Abstract Algebra*, 3rd edition, Blackie Publishing.

## Appendix

**Table 1. 2-ary Necklace Series with actual observations.**

| n | a(n) | observed |
|---|---|---|
| 1 | 1 | |
| 2 | 2 | 2 |
| 3 | 2 | |
| 4 | 3 | 3 |
| 5 | 3 | |
| 6 | 5 | 5 |
| 7 | 5 | |
| 8 | 8 | 8 |
| 9 | 10 | |
| 10 | 15 | 15 |
| 11 | 19 | |
| 12 | 31 | 31 |
| 13 | 41 | |
| 14 | 64 | 64 |
| 15 | 94 | |
| 16 | 143 | 143 |
| 17 | 211 | |
| 18 | 329 | |
| 19 | 493 | |
| 20 | 766 | |
| 21 | 1170 | |
| 22 | 1811 | |
| 23 | 2787 | |
| 24 | 4341 | |
| 25 | 6713 | |
| 26 | 10462 | |
| 27 | 16274 | |
| 28 | 25415 | |
| 29 | 39651 | |
| 30 | 62075 | |
| 31 | 97109 | |
| 32 | 152288 | |
| 33 | 238838 | |
| 34 | 375167 | |
| 35 | 589527 | |
| 36 | 927555 | |
| 37 | 1459961 | |
| 38 | 2300348 | |
| 39 | 3626242 | |
| 40 | 5721045 | |
| 41 | 9030451 | |
| 42 | 14264309 | |
| 43 | 22542397 | |

## Sensor fusion operations on $g^6$ groups: group transformations

$$\Phi_0 : G_1^6 \to G_1^6$$
$$\Phi_1 : G_1^6 \to G_5^6$$
$$\Phi_2 : G_1^6 \to G_5^6$$
$$\Phi_3 : G_1^6 \to G_9^6$$
$$\Phi_4 : G_1^6 \to G_5^6$$
$$\Phi_5 : G_1^6 \to G_5^6$$
$$\Phi_0 : G_5^6 \to G_5^6$$
$$\Phi_1 : G_5^6 \to G_{21}^6$$
$$\Phi_2 : G_5^6 \to G_5^6$$
$$\Phi_3 : G_5^6 \to G_9^6$$
$$\Phi_4 : G_5^6 \to G_{21}^6$$
$$\Phi_5 : G_5^6 \to G_9^6$$
$$\Phi_0 : G_9^6 \to G_9^6$$
$$\Phi_1 : G_9^6 \to G_{21}^6$$
$$\Phi_2 : G_9^6 \to G_9^6$$
$$\Phi_3 : G_9^6 \to G_9^6$$
$$\Phi_4 : G_9^6 \to G_{21}^6$$
$$\Phi_5 : G_9^6 \to G_{21}^6$$
$$\Phi_a : G_{21}^6 \to G_{21}^6$$

$$\Phi_{01} : G_1^6 \rightarrow G_5^6$$
$$\Phi_{12} : G_1^6 \rightarrow G_5^6$$
$$\Phi_{23} : G_1^6 \rightarrow G_9^6$$
$$\Phi_{34} : G_1^6 \rightarrow G_9^6$$
$$\Phi_{45} : G_1^6 \rightarrow G_5^6$$
$$\Phi_{01} : G_5^6 \rightarrow G_5^6$$
$$\Phi_{12} : G_5^6 \rightarrow G_5^6$$
$$\Phi_{23} : G_5^6 \rightarrow G_9^6$$
$$\Phi_{34} : G_5^6 \rightarrow G_{21}^6$$
$$\Phi_{45} : G_5^6 \rightarrow G_{21}^6$$
$$\Phi_{ab} : G_9^6 \rightarrow G_{21}^6$$
$$\Phi_{ab} : G_{21}^6 \rightarrow G_{21}^6$$
$$\Phi_{abc} : G_1^6 \rightarrow G_{21}^6$$
$$\Phi_{abc} : G_5^6 \rightarrow G_{21}^6$$
$$\Phi_{abc} : G_9^6 \rightarrow G_{21}^6$$
$$\Phi_{abc} : G_{21}^6 \rightarrow G_{21}^6$$
$$\Phi_{abcd} : G_1^6 \rightarrow G_{21}^6$$
$$\Phi_{abcd} : G_5^6 \rightarrow G_{21}^6$$
$$\Phi_{abcd} : G_9^6 \rightarrow G_{21}^6$$
$$\Phi_{abcd} : G_{21}^6 \rightarrow G_{21}^6$$
$$a, b, c, d \in \{0,1,2,3,4,5\}$$
$$a \neq b \neq c \neq d$$

**Sensor fusion operations on $g^8$ groups: group transformations**

$$\Phi_0 : G_1^8 \rightarrow G_1^8$$
$$\Phi_1 : G_1^8 \rightarrow G_5^8$$
$$\Phi_2 : G_1^8 \rightarrow G_5^8$$
$$\Phi_3 : G_1^8 \rightarrow G_9^8$$
$$\Phi_4 : G_1^8 \rightarrow G_{17}^8$$
$$\Phi_5 : G_1^8 \rightarrow G_9^8$$
$$\Phi_6 : G_1^8 \rightarrow G_5^8$$
$$\Phi_7 : G_1^8 \rightarrow G_5^8$$
$$\Phi_0 : G_5^8 \rightarrow G_5^8$$
$$\Phi_1 : G_5^8 \rightarrow G_5^8$$
$$\Phi_2 : G_5^8 \rightarrow G_5^8$$
$$\Phi_3 : G_5^8 \rightarrow G_{21}^8$$
$$\Phi_4 : G_5^8 \rightarrow G_{21}^8$$
$$\Phi_5 : G_5^8 \rightarrow G_{37}^8$$
$$\Phi_6 : G_5^8 \rightarrow G_{21}^8$$
$$\Phi_7 : G_5^8 \rightarrow G_{21}^8$$
$$\Phi_0 : G_9^8 \rightarrow G_9^8$$
$$\Phi_1 : G_9^8 \rightarrow G_{21}^8$$
$$\Phi_2 : G_9^8 \rightarrow G_{21}^8$$
$$\Phi_3 : G_9^8 \rightarrow G_9^8$$
$$\Phi_4 : G_9^8 \rightarrow G_{37}^8$$
$$\Phi_5 : G_9^8 \rightarrow G_{37}^8$$
$$\Phi_6 : G_9^8 \rightarrow G_{37}^8$$
$$\Phi_7 : G_9^8 \rightarrow G_{37}^8$$
$$\Phi_0 : G_{17}^8 \rightarrow G_{17}^8$$
$$\Phi_1 : G_{17}^8 \rightarrow G_{21}^8$$
$$\Phi_2 : G_{17}^8 \rightarrow G_{37}^8$$
$$\Phi_3 : G_{17}^8 \rightarrow G_{37}^8$$
$$\Phi_4 : G_{17}^8 \rightarrow G_{17}^8$$
$$\Phi_5 : G_{17}^8 \rightarrow G_{37}^8$$
$$\Phi_6 : G_{17}^8 \rightarrow G_{37}^8$$
$$\Phi_7 : G_{17}^8 \rightarrow G_{37}^8$$

$$\Phi_0 : G_{21}^8 \to G_{21}^8$$
$$\Phi_1 : G_{21}^8 \to G_{85}^8$$
$$\Phi_2 : G_{21}^8 \to G_{21}^8$$
$$\Phi_3 : G_{21}^8 \to G_{85}^8$$
$$\Phi_4 : G_{21}^8 \to G_{21}^8$$
$$\Phi_5 : G_{21}^8 \to G_{85}^8$$
$$\Phi_6 : G_{21}^8 \to G_{85}^8$$
$$\Phi_7 : G_{21}^8 \to G_{85}^8$$
$$\Phi_0 : G_{37}^8 \to G_{37}^8$$
$$\Phi_1 : G_{37}^8 \to G_{85}^8$$
$$\Phi_2 : G_{37}^8 \to G_{37}^8$$
$$\Phi_3 : G_{37}^8 \to G_{85}^8$$
$$\Phi_4 : G_{37}^8 \to G_{85}^8$$
$$\Phi_5 : G_{37}^8 \to G_{37}^8$$
$$\Phi_6 : G_{37}^8 \to G_{85}^8$$
$$\Phi_7 : G_{37}^8 \to G_{85}^8$$
$$\Phi_{01} : G_1^8 \to G_5^8$$
$$\Phi_{12} : G_1^8 \to G_9^8$$
$$\Phi_{23} : G_1^8 \to G_{21}^8$$
$$\Phi_{34} : G_1^8 \to G_{37}^8$$
$$\Phi_{45} : G_1^8 \to G_{37}^8$$
$$\Phi_{56} : G_1^8 \to G_{21}^8$$
$$\Phi_{67} : G_1^8 \to G_{37}^8$$
$$\Phi_{01} : G_5^8 \to G_{21}^8$$
$$\Phi_{12} : G_5^8 \to G_{21}^8$$
$$\Phi_{23} : G_5^8 \to G_{37}^8$$
$$\Phi_{34} : G_5^8 \to G_{85}^8$$
$$\Phi_{45} : G_5^8 \to G_{85}^8$$
$$\Phi_{56} : G_5^8 \to G_{85}^8$$
$$\Phi_{67} : G_5^8 \to G_{85}^8$$

$$\Phi_{01}: G_9^8 \to G_{21}^8$$
$$\Phi_{12}: G_9^8 \to G_{85}^8$$
$$\Phi_{23}: G_9^8 \to G_{21}^8$$
$$\Phi_{34}: G_9^8 \to G_{85}^8$$
$$\Phi_{45}: G_9^8 \to G_{85}^8$$
$$\Phi_{56}: G_9^8 \to G_{85}^8$$
$$\Phi_{67}: G_9^8 \to G_{85}^8$$
$$\Phi_{01}: G_{17}^8 \to G_{37}^8$$
$$\Phi_{12}: G_{17}^8 \to G_{85}^8$$
$$\Phi_{23}: G_{17}^8 \to G_{85}^8$$
$$\Phi_{34}: G_{17}^8 \to G_{37}^8$$
$$\Phi_{45}: G_{17}^8 \to G_{37}^8$$
$$\Phi_{56}: G_{17}^8 \to G_{85}^8$$
$$\Phi_{67}: G_{17}^8 \to G_{85}^8$$
$$\Phi_{012}: G_1^8 \to G_{37}^8$$
$$\Phi_{123}: G_1^8 \to G_{85}^8$$
$$\Phi_{234}: G_1^8 \to G_{85}^8$$
$$\Phi_{345}: G_1^8 \to G_{85}^8$$
$$\Phi_{456}: G_1^8 \to G_{85}^8$$
$$\Phi_{567}: G_1^8 \to G_{85}^8$$
$$\Phi_{\alpha}: G_{85}^8 \to G_{85}^8 \quad 8-ways$$
$$\Phi_{\alpha\beta}: G_{21}^8 \to G_{85}^8 \quad 7-ways$$
$$\Phi_{\alpha\beta}: G_{37}^8 \to G_{85}^8 \quad 7-ways$$
$$\Phi_{\alpha\beta}: G_{85}^8 \to G_{85}^8 \quad 7-ways$$
$$\Phi_{\alpha\beta\chi}: G_9^8 \to G_{85}^8 \quad 6-ways$$
$$\Phi_{\alpha\beta\chi}: G_{17}^8 \to G_{85}^8 \quad 6-ways$$
$$\Phi_{\alpha\beta\chi\delta}: G_i^8 \to G_{85}^8 \quad 35-ways$$
$$\Phi_{\alpha\beta\chi\delta\varepsilon}: G_i^8 \to G_{85}^8 \quad 28-ways$$

where

$$\alpha, \beta, \chi, \delta, \varepsilon \in \{0,1,2,3,4,5)$$
$$i \in \{\underline{1}, \underline{5}, \underline{9}, \underline{17}, \underline{21}, \underline{37}, \underline{85}\}$$

## 10-node machine Group Operations

```
1-->2-->4-->8-->16-->32-->64-->128-->256-->512-->: 10-cycle
state: 1 = 00000000000000000000000000000001
5-->10-->20-->40-->80-->160-->320-->640-->257-->514-->: 10-cycle
state: 5 = 00000000000000000000000000000101
9-->18-->36-->72-->144-->288-->576-->129-->258-->516-->: 10-cycle
state: 9 = 00000000000000000000000000001001
17-->34-->68-->136-->272-->544-->65-->130-->260-->520-->: 10-cycle
state: 17 = 00000000000000000000000000010001
21-->42-->84-->168-->336-->672-->321-->642-->261-->522-->: 10-cycle
state: 21 = 00000000000000000000000000010101
33-->66-->132-->264-->528-->33-->66-->132-->264-->528-->: 5-cycle
state: 33 = 00000000000000000000000000100001
37-->74-->148-->296-->592-->161-->322-->644-->265-->530-->: 10-cycle
state: 37 = 00000000000000000000000000100101
41-->82-->164-->328-->656-->289-->578-->133-->266-->532-->: 10-cycle
state: 41 = 00000000000000000000000000101001
69-->138-->276-->552-->81-->162-->324-->648-->273-->546-->: 10-cycle
state: 69 = 00000000000000000000000001000101
73-->146-->292-->584-->145-->290-->580-->137-->274-->548-->: 10-cycle
state: 73 = 00000000000000000000000001001001
85-->170-->340-->680-->337-->674-->325-->650-->277-->554-->: 10-cycle
state: 85 = 00000000000000000000000001010101
149-->298-->596-->169-->338-->676-->329-->658-->293-->586-->: 10-cycle
state: 149 = 00000000000000000000000010010101
165-->330-->660-->297-->594-->165-->330-->660-->297-->594-->: 5-cycle
state: 165 = 00000000000000000000000010100101
341-->682-->341-->682-->341-->682-->341-->682-->341-->682-->: 2-cycle
state: 341 = 00000000000000000000000101010101
```

## 12-node machine Group Operations

```
1-->2-->4-->8-->16-->32-->64-->128-->256-->512-->1024-->2048-->: 12-
cycle
state: 1 = 00000000000000000000000000000001
5-->10-->20-->40-->80-->160-->320-->640-->1280-->2560-->1025-->2050-->:
12-cycle
state: 5 = 00000000000000000000000000000101
9-->18-->36-->72-->144-->288-->576-->1152-->2304-->513-->1026-->2052--
>: 12-cycle
state: 9 = 00000000000000000000000000001001
17-->34-->68-->136-->272-->544-->1088-->2176-->257-->514-->1028-->2056-
->: 12-cycle
state: 17 = 00000000000000000000000000010001
21-->42-->84-->168-->336-->672-->1344-->2688-->1281-->2562-->1029--
>2058-->: 12-cycle
state: 21 = 00000000000000000000000000010101
33-->66-->132-->264-->528-->1056-->2112-->129-->258-->516-->1032--
>2064-->: 12-cycle
state: 33 = 00000000000000000000000000100001
37-->74-->148-->296-->592-->1184-->2368-->641-->1282-->2564-->1033--
>2066-->: 12-cycle
state: 37 = 00000000000000000000000000100101
```

```
41-->82-->164-->328-->656-->1312-->2624-->1153-->2306-->517-->1034--
>2068-->: 12-cycle
state: 41 = 00000000000000000000000000101001
65-->130-->260-->520-->1040-->2080-->65-->130-->260-->520-->1040--
>2080-->: 6-cycle
state: 65 = 00000000000000000000000001000001
69-->138-->276-->552-->1104-->2208-->321-->642-->1284-->2568-->1041--
>2082-->: 12-cycle
state: 69 = 00000000000000000000000001000101
73-->146-->292-->584-->1168-->2336-->577-->1154-->2308-->521-->1042--
>2084-->: 12-cycle
state: 73 = 00000000000000000000000001001001
81-->162-->324-->648-->1296-->2592-->1089-->2178-->261-->522-->1044--
>2088-->: 12-cycle
state: 81 = 00000000000000000000000001010001
85-->170-->340-->680-->1360-->2720-->1345-->2690-->1285-->2570-->1045--
>2090-->: 12-cycle
state: 85 = 00000000000000000000000001010101
133-->266-->532-->1064-->2128-->161-->322-->644-->1288-->2576-->1057--
>2114-->: 12-cycle
state: 133 = 00000000000000000000000010000101
137-->274-->548-->1096-->2192-->289-->578-->1156-->2312-->529-->1058--
>2116-->: 12-cycle
state: 137 = 00000000000000000000000010001001
145-->290-->580-->1160-->2320-->545-->1090-->2180-->265-->530-->1060--
>2120-->: 12-cycle
state: 145 = 00000000000000000000000010010001
149-->298-->596-->1192-->2384-->673-->1346-->2692-->1289-->2578-->1061-
->2122-->: 12-cycle
state: 149 = 00000000000000000000000010010101
165-->330-->660-->1320-->2640-->1185-->2370-->645-->1290-->2580-->1065-
->2130-->: 12-cycle
state: 165 = 00000000000000000000000010100101
169-->338-->676-->1352-->2704-->1313-->2626-->1157-->2314-->533-->1066-
->2132-->: 12-cycle
state: 169 = 00000000000000000000000010101001
273-->546-->1092-->2184-->273-->546-->1092-->2184-->273-->546-->1092--
>2184-->: 4-cycle
state: 273 = 00000000000000000000000100010001
277-->554-->1108-->2216-->337-->674-->1348-->2696-->1297-->2594-->1093-
->2186-->: 12-cycle
state: 277 = 00000000000000000000000100010101
293-->586-->1172-->2344-->593-->1186-->2372-->649-->1298-->2596-->1097-
->2194-->: 12-cycle
state: 293 = 00000000000000000000000100100101
297-->594-->1188-->2376-->657-->1314-->2628-->1161-->2322-->549-->1098-
->2196-->: 12-cycle
state: 297 = 00000000000000000000000100101001
325-->650-->1300-->2600-->1105-->2210-->325-->650-->1300-->2600-->1105-
->2210-->: 6-cycle
state: 325 = 00000000000000000000000101000101
329-->658-->1316-->2632-->1169-->2338-->581-->1162-->2324-->553-->1106-
->2212-->: 12-cycle
state: 329 = 00000000000000000000000101001001
341-->682-->1364-->2728-->1361-->2722-->1349-->2698-->1301-->2602--
>1109-->2218-->: 12-cycle
state: 341 = 00000000000000000000000101010101
```

```
585-->1170-->2340-->585-->1170-->2340-->585-->1170-->2340-->585-->1170-
->2340-->: 3-cycle
state: 585 = 00000000000000000000001001001001
597-->1194-->2388-->681-->1362-->2724-->1353-->2706-->1317-->2634--
>1173-->2346-->: 12-cycle
state: 597 = 00000000000000000000001001010101
661-->1322-->2644-->1193-->2386-->677-->1354-->2708-->1321-->2642--
>1189-->2378-->: 12-cycle
state: 661 = 00000000000000000000001010010101
1365-->2730-->1365-->2730-->1365-->2730-->1365-->2730-->1365-->2730--
>1365-->2730-->: 2-cycle
state: 1365 = 00000000000000000000010101010101
```

## 14-node machine Group Operations

```
1-->2-->4-->8-->16-->32-->64-->128-->256-->512-->1024-->2048-->4096--
>8192-->: 14-cycle
state: 1 = 00000000000000000000000000000001
5-->10-->20-->40-->80-->160-->320-->640-->1280-->2560-->5120-->10240--
>4097-->8194-->: 14-cycle
state: 5 = 00000000000000000000000000000101
9-->18-->36-->72-->144-->288-->576-->1152-->2304-->4608-->9216-->2049--
>4098-->8196-->: 14-cycle
state: 9 = 00000000000000000000000000001001
17-->34-->68-->136-->272-->544-->1088-->2176-->4352-->8704-->1025--
>2050-->4100-->8200-->: 14-cycle
state: 17 = 00000000000000000000000000010001
21-->42-->84-->168-->336-->672-->1344-->2688-->5376-->10752-->5121--
>10242-->4101-->8202-->: 14-cycle
state: 21 = 00000000000000000000000000010101
33-->66-->132-->264-->528-->1056-->2112-->4224-->8448-->513-->1026--
>2052-->4104-->8208-->: 14-cycle
state: 33 = 00000000000000000000000000100001
37-->74-->148-->296-->592-->1184-->2368-->4736-->9472-->2561-->5122--
>10244-->4105-->8210-->: 14-cycle
state: 37 = 00000000000000000000000000100101
41-->82-->164-->328-->656-->1312-->2624-->5248-->10496-->4609-->9218--
>2053-->4106-->8212-->: 14-cycle
state: 41 = 00000000000000000000000000101001
65-->130-->260-->520-->1040-->2080-->4160-->8320-->257-->514-->1028--
>2056-->4112-->8224-->: 14-cycle
state: 65 = 00000000000000000000000001000001
69-->138-->276-->552-->1104-->2208-->4416-->8832-->1281-->2562-->5124--
>10248-->4113-->8226-->: 14-cycle
state: 69 = 00000000000000000000000001000101
73-->146-->292-->584-->1168-->2336-->4672-->9344-->2305-->4610-->9220--
>2057-->4114-->8228-->: 14-cycle
state: 73 = 00000000000000000000000001001001
81-->162-->324-->648-->1296-->2592-->5184-->10368-->4353-->8706-->1029-
->2058-->4116-->8232-->: 14-cycle
state: 81 = 00000000000000000000000001010001
85-->170-->340-->680-->1360-->2720-->5440-->10880-->5377-->10754--
>5125-->10250-->4117-->8234-->: 14-cycle
state: 85 = 00000000000000000000000001010101
```

129-->258-->516-->1032-->2064-->4128-->8256-->129-->258-->516-->1032--
>2064-->4128-->8256-->: 7-cycle
state: 129 = 00000000000000000000000010000001
133-->266-->532-->1064-->2128-->4256-->8512-->641-->1282-->2564-->5128-
->10256-->4129-->8258-->: 14-cycle
state: 133 = 00000000000000000000000010000101
137-->274-->548-->1096-->2192-->4384-->8768-->1153-->2306-->4612--
>9224-->2065-->4130-->8260-->: 14-cycle
state: 137 = 00000000000000000000000010001001
145-->290-->580-->1160-->2320-->4640-->9280-->2177-->4354-->8708--
>1033-->2066-->4132-->8264-->: 14-cycle
state: 145 = 00000000000000000000000010010001
149-->298-->596-->1192-->2384-->4768-->9536-->2689-->5378-->10756--
>5129-->10258-->4133-->8266-->: 14-cycle
state: 149 = 00000000000000000000000010010101
161-->322-->644-->1288-->2576-->5152-->10304-->4225-->8450-->517--
>1034-->2068-->4136-->8272-->: 14-cycle
state: 161 = 00000000000000000000000010100001
165-->330-->660-->1320-->2640-->5280-->10560-->4737-->9474-->2565--
>5130-->10260-->4137-->8274-->: 14-cycle
state: 165 = 00000000000000000000000010100101
169-->338-->676-->1352-->2704-->5408-->10816-->5249-->10498-->4613--
>9226-->2069-->4138-->8276-->: 14-cycle
state: 169 = 00000000000000000000000010101001
261-->522-->1044-->2088-->4176-->8352-->321-->642-->1284-->2568-->5136-
->10272-->4161-->8322-->: 14-cycle
state: 261 = 00000000000000000000000100000101
265-->530-->1060-->2120-->4240-->8480-->577-->1154-->2308-->4616--
>9232-->2081-->4162-->8324-->: 14-cycle
state: 265 = 00000000000000000000000100001001
273-->546-->1092-->2184-->4368-->8736-->1089-->2178-->4356-->8712--
>1041-->2082-->4164-->8328-->: 14-cycle
state: 273 = 00000000000000000000000100010001
277-->554-->1108-->2216-->4432-->8864-->1345-->2690-->5380-->10760--
>5137-->10274-->4165-->8330-->: 14-cycle
state: 277 = 00000000000000000000000100010101
289-->578-->1156-->2312-->4624-->9248-->2113-->4226-->8452-->521--
>1042-->2084-->4168-->8336-->: 14-cycle
state: 289 = 00000000000000000000000100100001
293-->586-->1172-->2344-->4688-->9376-->2369-->4738-->9476-->2569--
>5138-->10276-->4169-->8338-->: 14-cycle
state: 293 = 00000000000000000000000100100101
297-->594-->1188-->2376-->4752-->9504-->2625-->5250-->10500-->4617--
>9234-->2085-->4170-->8340-->: 14-cycle
state: 297 = 00000000000000000000000100101001
325-->650-->1300-->2600-->5200-->10400-->4417-->8834-->1285-->2570--
>5140-->10280-->4177-->8354-->: 14-cycle
state: 325 = 00000000000000000000000101000101
329-->658-->1316-->2632-->5264-->10528-->4673-->9346-->2309-->4618--
>9236-->2089-->4178-->8356-->: 14-cycle
state: 329 = 00000000000000000000000101001001
337-->674-->1348-->2696-->5392-->10784-->5185-->10370-->4357-->8714--
>1045-->2090-->4180-->8360-->: 14-cycle
state: 337 = 00000000000000000000000101010001
341-->682-->1364-->2728-->5456-->10912-->5441-->10882-->5381-->10762--
>5141-->10282-->4181-->8362-->: 14-cycle
state: 341 = 00000000000000000000000101010101

```
529-->1058-->2116-->4232-->8464-->545-->1090-->2180-->4360-->8720--
>1057-->2114-->4228-->8456-->: 14-cycle
state: 529 = 00000000000000000000001000010001
533-->1066-->2132-->4264-->8528-->673-->1346-->2692-->5384-->10768--
>5153-->10306-->4229-->8458-->: 14-cycle
state: 533 = 00000000000000000000001000010101
549-->1098-->2196-->4392-->8784-->1185-->2370-->4740-->9480-->2577--
>5154-->10308-->4233-->8466-->: 14-cycle
state: 549 = 00000000000000000000001000100101
553-->1106-->2212-->4424-->8848-->1313-->2626-->5252-->10504-->4625--
>9250-->2117-->4234-->8468-->: 14-cycle
state: 553 = 00000000000000000000001000101001
581-->1162-->2324-->4648-->9296-->2209-->4418-->8836-->1289-->2578--
>5156-->10312-->4241-->8482-->: 14-cycle
state: 581 = 00000000000000000000001001000101
585-->1170-->2340-->4680-->9360-->2337-->4674-->9348-->2313-->4626--
>9252-->2121-->4242-->8484-->: 14-cycle
state: 585 = 00000000000000000000001001001001
593-->1186-->2372-->4744-->9488-->2593-->5186-->10372-->4361-->8722--
>1061-->2122-->4244-->8488-->: 14-cycle
state: 593 = 00000000000000000000001001010001
597-->1194-->2388-->4776-->9552-->2721-->5442-->10884-->5385-->10770--
>5157-->10314-->4245-->8490-->: 14-cycle
state: 597 = 00000000000000000000001001010101
645-->1290-->2580-->5160-->10320-->4257-->8514-->645-->1290-->2580--
>5160-->10320-->4257-->8514-->: 7-cycle
state: 645 = 00000000000000000000001010000101
649-->1298-->2596-->5192-->10384-->4385-->8770-->1157-->2314-->4628--
>9256-->2129-->4258-->8516-->: 14-cycle
state: 649 = 00000000000000000000001010001001
657-->1314-->2628-->5256-->10512-->4641-->9282-->2181-->4362-->8724--
>1065-->2130-->4260-->8520-->: 14-cycle
state: 657 = 00000000000000000000001010010001
661-->1322-->2644-->5288-->10576-->4769-->9538-->2693-->5386-->10772--
>5161-->10322-->4261-->8522-->: 14-cycle
state: 661 = 00000000000000000000001010010101
677-->1354-->2708-->5416-->10832-->5281-->10562-->4741-->9482-->2581--
>5162-->10324-->4265-->8530-->: 14-cycle
state: 677 = 00000000000000000000001010100101
681-->1362-->2724-->5448-->10896-->5409-->10818-->5253-->10506-->4629--
>9258-->2133-->4266-->8532-->: 14-cycle
state: 681 = 00000000000000000000001010101001
1093-->2186-->4372-->8744-->1105-->2210-->4420-->8840-->1297-->2594--
>5188-->10376-->4369-->8738-->: 14-cycle
state: 1093 = 00000000000000000000010001000101
1097-->2194-->4388-->8776-->1169-->2338-->4676-->9352-->2321-->4642--
>9284-->2185-->4370-->8740-->: 14-cycle
state: 1097 = 00000000000000000000010001001001
1109-->2218-->4436-->8872-->1361-->2722-->5444-->10888-->5393-->10786--
>5189-->10378-->4373-->8746-->: 14-cycle
state: 1109 = 00000000000000000000010001010101
1161-->2322-->4644-->9288-->2193-->4386-->8772-->1161-->2322-->4644--
>9288-->2193-->4386-->8772-->: 7-cycle
state: 1161 = 00000000000000000000010010001001
1173-->2346-->4692-->9384-->2385-->4770-->9540-->2697-->5394-->10788--
>5193-->10386-->4389-->8778-->: 14-cycle
state: 1173 = 00000000000000000000010010010101
```

1189-->2378-->4756-->9512-->2641-->5282-->10564-->4745-->9490-->2597--
>5194-->10388-->4393-->8786-->: 14-cycle
state: 1189 = 00000000000000000000010010100101
1193-->2386-->4772-->9544-->2705-->5410-->10820-->5257-->10514-->4645--
>9290-->2197-->4394-->8788-->: 14-cycle
state: 1193 = 00000000000000000000010010101001
1301-->2602-->5204-->10408-->4433-->8866-->1349-->2698-->5396-->10792--
>5201-->10402-->4421-->8842-->: 14-cycle
state: 1301 = 00000000000000000000010100010101
1317-->2634-->5268-->10536-->4689-->9378-->2373-->4746-->9492-->2601--
>5202-->10404-->4425-->8850-->: 14-cycle
state: 1317 = 00000000000000000000010100100101
1321-->2642-->5284-->10568-->4753-->9506-->2629-->5258-->10516-->4649--
>9298-->2213-->4426-->8852-->: 14-cycle
state: 1321 = 00000000000000000000010100101001
1353-->2706-->5412-->10824-->5265-->10530-->4677-->9354-->2325-->4650--
>9300-->2217-->4434-->8868-->: 14-cycle
state: 1353 = 00000000000000000000010101001001
1365-->2730-->5460-->10920-->5457-->10914-->5445-->10890-->5397--
>10794-->5205-->10410-->4437-->8874-->: 14-cycle
state: 1365 = 00000000000000000000010101010101
2341-->4682-->9364-->2345-->4690-->9380-->2377-->4754-->9508-->2633--
>5266-->10532-->4681-->9362-->: 14-cycle
state: 2341 = 00000000000000000000100100100101
2389-->4778-->9556-->2729-->5458-->10916-->5449-->10898-->5413-->10826-
->5269-->10538-->4693-->9386-->: 14-cycle
state: 2389 = 00000000000000000000100101010101
2645-->5290-->10580-->4777-->9554-->2725-->5450-->10900-->5417-->10834-
->5285-->10570-->4757-->9514-->: 14-cycle
state: 2645 = 00000000000000000000101001010101
2709-->5418-->10836-->5289-->10578-->4773-->9546-->2709-->5418-->10836-
->5289-->10578-->4773-->9546-->: 7-cycle
state: 2709 = 00000000000000000000101010010101
5461-->10922-->5461-->10922-->5461-->10922-->5461-->10922-->5461--
>10922-->5461-->10922-->5461-->10922-->: 2-cycle
state: 5461 = 00000000000000000001010101010101

## 16-node machine Group Operations

1-->2-->4-->8-->16-->32-->64-->128-->256-->512-->1024-->2048-->4096--
>8192-->16384-->32768-->: 16-cycle
state: 1 = 00000000000000000000000000000001
5-->10-->20-->40-->80-->160-->320-->640-->1280-->2560-->5120-->10240--
>20480-->40960-->16385-->32770-->: 16-cycle
state: 5 = 00000000000000000000000000000101
9-->18-->36-->72-->144-->288-->576-->1152-->2304-->4608-->9216-->18432-
->36864-->8193-->16386-->32772-->: 16-cycle
state: 9 = 00000000000000000000000000001001
17-->34-->68-->136-->272-->544-->1088-->2176-->4352-->8704-->17408--
>34816-->4097-->8194-->16388-->32776-->: 16-cycle
state: 17 = 00000000000000000000000000010001
21-->42-->84-->168-->336-->672-->1344-->2688-->5376-->10752-->21504--
>43008-->20481-->40962-->16389-->32778-->: 16-cycle
state: 21 = 00000000000000000000000000010101

```
33-->66-->132-->264-->528-->1056-->2112-->4224-->8448-->16896-->33792--
>2049-->4098-->8196-->16392-->32784-->: 16-cycle
state: 33 = 00000000000000000000000000100001
37-->74-->148-->296-->592-->1184-->2368-->4736-->9472-->18944-->37888--
>10241-->20482-->40964-->16393-->32786-->: 16-cycle
state: 37 = 00000000000000000000000000100101
41-->82-->164-->328-->656-->1312-->2624-->5248-->10496-->20992-->41984-
->18433-->36866-->8197-->16394-->32788-->: 16-cycle
state: 41 = 00000000000000000000000000101001
65-->130-->260-->520-->1040-->2080-->4160-->8320-->16640-->33280--
>1025-->2050-->4100-->8200-->16400-->32800-->: 16-cycle
state: 65 = 00000000000000000000000001000001
69-->138-->276-->552-->1104-->2208-->4416-->8832-->17664-->35328--
>5121-->10242-->20484-->40968-->16401-->32802-->: 16-cycle
state: 69 = 00000000000000000000000001000101
73-->146-->292-->584-->1168-->2336-->4672-->9344-->18688-->37376--
>9217-->18434-->36868-->8201-->16402-->32804-->: 16-cycle
state: 73 = 00000000000000000000000001001001
81-->162-->324-->648-->1296-->2592-->5184-->10368-->20736-->41472--
>17409-->34818-->4101-->8202-->16404-->32808-->: 16-cycle
state: 81 = 00000000000000000000000001010001
85-->170-->340-->680-->1360-->2720-->5440-->10880-->21760-->43520--
>21505-->43010-->20485-->40970-->16405-->32810-->: 16-cycle
state: 85 = 00000000000000000000000001010101
129-->258-->516-->1032-->2064-->4128-->8256-->16512-->33024-->513--
>1026-->2052-->4104-->8208-->16416-->32832-->: 16-cycle
state: 129 = 00000000000000000000000010000001
133-->266-->532-->1064-->2128-->4256-->8512-->17024-->34048-->2561--
>5122-->10244-->20488-->40976-->16417-->32834-->: 16-cycle
state: 133 = 00000000000000000000000010000101
137-->274-->548-->1096-->2192-->4384-->8768-->17536-->35072-->4609--
>9218-->18436-->36872-->8209-->16418-->32836-->: 16-cycle
state: 137 = 00000000000000000000000010001001
145-->290-->580-->1160-->2320-->4640-->9280-->18560-->37120-->8705--
>17410-->34820-->4105-->8210-->16420-->32840-->: 16-cycle
state: 145 = 00000000000000000000000010010001
149-->298-->596-->1192-->2384-->4768-->9536-->19072-->38144-->10753--
>21506-->43012-->20489-->40978-->16421-->32842-->: 16-cycle
state: 149 = 00000000000000000000000010010101
161-->322-->644-->1288-->2576-->5152-->10304-->20608-->41216-->16897--
>33794-->2053-->4106-->8212-->16424-->32848-->: 16-cycle
state: 161 = 00000000000000000000000010100001
165-->330-->660-->1320-->2640-->5280-->10560-->21120-->42240-->18945--
>37890-->10245-->20490-->40980-->16425-->32850-->: 16-cycle
state: 165 = 00000000000000000000000010100101
169-->338-->676-->1352-->2704-->5408-->10816-->21632-->43264-->20993--
>41986-->18437-->36874-->8213-->16426-->32852-->: 16-cycle
state: 169 = 00000000000000000000000010101001
257-->514-->1028-->2056-->4112-->8224-->16448-->32896-->257-->514--
>1028-->2056-->4112-->8224-->16448-->32896-->: 8-cycle
state: 257 = 00000000000000000000000100000001
261-->522-->1044-->2088-->4176-->8352-->16704-->33408-->1281-->2562--
>5124-->10248-->20496-->40992-->16449-->32898-->: 16-cycle
state: 261 = 00000000000000000000000100000101
265-->530-->1060-->2120-->4240-->8480-->16960-->33920-->2305-->4610--
>9220-->18440-->36880-->8225-->16450-->32900-->: 16-cycle
state: 265 = 00000000000000000000000100001001
```

273-->546-->1092-->2184-->4368-->8736-->17472-->34944-->4353-->8706--
>17412-->34824-->4113-->8226-->16452-->32904-->: 16-cycle
state: 273 = 00000000000000000000000100010001
277-->554-->1108-->2216-->4432-->8864-->17728-->35456-->5377-->10754--
>21508-->43016-->20497-->40994-->16453-->32906-->: 16-cycle
state: 277 = 00000000000000000000000100010101
289-->578-->1156-->2312-->4624-->9248-->18496-->36992-->8449-->16898--
>33796-->2057-->4114-->8228-->16456-->32912-->: 16-cycle
state: 289 = 00000000000000000000000100100001
293-->586-->1172-->2344-->4688-->9376-->18752-->37504-->9473-->18946--
>37892-->10249-->20498-->40996-->16457-->32914-->: 16-cycle
state: 293 = 00000000000000000000000100100101
297-->594-->1188-->2376-->4752-->9504-->19008-->38016-->10497-->20994--
>41988-->18441-->36882-->8229-->16458-->32916-->: 16-cycle
state: 297 = 00000000000000000000000100101001
321-->642-->1284-->2568-->5136-->10272-->20544-->41088-->16641-->33282-
->1029-->2058-->4116-->8232-->16464-->32928-->: 16-cycle
state: 321 = 00000000000000000000000101000001
325-->650-->1300-->2600-->5200-->10400-->20800-->41600-->17665-->35330-
->5125-->10250-->20500-->41000-->16465-->32930-->: 16-cycle
state: 325 = 00000000000000000000000101000101
329-->658-->1316-->2632-->5264-->10528-->21056-->42112-->18689-->37378-
->9221-->18442-->36884-->8233-->16466-->32932-->: 16-cycle
state: 329 = 00000000000000000000000101001001
337-->674-->1348-->2696-->5392-->10784-->21568-->43136-->20737-->41474-
->17413-->34826-->4117-->8234-->16468-->32936-->: 16-cycle
state: 337 = 00000000000000000000000101010001
341-->682-->1364-->2728-->5456-->10912-->21824-->43648-->21761-->43522-
->21509-->43018-->20501-->41002-->16469-->32938-->: 16-cycle
state: 341 = 00000000000000000000000101010101
517-->1034-->2068-->4136-->8272-->16544-->33088-->641-->1282-->2564--
>5128-->10256-->20512-->41024-->16513-->33026-->: 16-cycle
state: 517 = 00000000000000000000001000000101
521-->1042-->2084-->4168-->8336-->16672-->33344-->1153-->2306-->4612--
>9224-->18448-->36896-->8257-->16514-->33028-->: 16-cycle
state: 521 = 00000000000000000000001000001001
529-->1058-->2116-->4232-->8464-->16928-->33856-->2177-->4354-->8708--
>17416-->34832-->4129-->8258-->16516-->33032-->: 16-cycle
state: 529 = 00000000000000000000001000010001
533-->1066-->2132-->4264-->8528-->17056-->34112-->2689-->5378-->10756--
>21512-->43024-->20513-->41026-->16517-->33034-->: 16-cycle
state: 533 = 00000000000000000000001000010101
545-->1090-->2180-->4360-->8720-->17440-->34880-->4225-->8450-->16900--
>33800-->2065-->4130-->8260-->16520-->33040-->: 16-cycle
state: 545 = 00000000000000000000001000100001
549-->1098-->2196-->4392-->8784-->17568-->35136-->4737-->9474-->18948--
>37896-->10257-->20514-->41028-->16521-->33042-->: 16-cycle
state: 549 = 00000000000000000000001000100101
553-->1106-->2212-->4424-->8848-->17696-->35392-->5249-->10498-->20996-
->41992-->18449-->36898-->8261-->16522-->33044-->: 16-cycle
state: 553 = 00000000000000000000001000101001
577-->1154-->2308-->4616-->9232-->18464-->36928-->8321-->16642-->33284-
->1033-->2066-->4132-->8264-->16528-->33056-->: 16-cycle
state: 577 = 00000000000000000000001001000001
581-->1162-->2324-->4648-->9296-->18592-->37184-->8833-->17666-->35332-
->5129-->10258-->20516-->41032-->16529-->33058-->: 16-cycle
state: 581 = 00000000000000000000001001000101

585-->1170-->2340-->4680-->9360-->18720-->37440-->9345-->18690-->37380-
->9225-->18450-->36900-->8265-->16530-->33060-->: 16-cycle
state: 585 = 00000000000000000000001001001001
593-->1186-->2372-->4744-->9488-->18976-->37952-->10369-->20738--
>41476-->17417-->34834-->4133-->8266-->16532-->33064-->: 16-cycle
state: 593 = 00000000000000000000001001010001
597-->1194-->2388-->4776-->9552-->19104-->38208-->10881-->21762--
>43524-->21513-->43026-->20517-->41034-->16533-->33066-->: 16-cycle
state: 597 = 00000000000000000000001001010101
645-->1290-->2580-->5160-->10320-->20640-->41280-->17025-->34050--
>2565-->5130-->10260-->20520-->41040-->16545-->33090-->: 16-cycle
state: 645 = 00000000000000000000001010000101
649-->1298-->2596-->5192-->10384-->20768-->41536-->17537-->35074--
>4613-->9226-->18452-->36904-->8273-->16546-->33092-->: 16-cycle
state: 649 = 00000000000000000000001010001001
657-->1314-->2628-->5256-->10512-->21024-->42048-->18561-->37122--
>8709-->17418-->34836-->4137-->8274-->16548-->33096-->: 16-cycle
state: 657 = 00000000000000000000001010010001
661-->1322-->2644-->5288-->10576-->21152-->42304-->19073-->38146--
>10757-->21514-->43028-->20521-->41042-->16549-->33098-->: 16-cycle
state: 661 = 00000000000000000000001010010101
673-->1346-->2692-->5384-->10768-->21536-->43072-->20609-->41218--
>16901-->33802-->2069-->4138-->8276-->16552-->33104-->: 16-cycle
state: 673 = 00000000000000000000001010100001
677-->1354-->2708-->5416-->10832-->21664-->43328-->21121-->42242--
>18949-->37898-->10261-->20522-->41044-->16553-->33106-->: 16-cycle
state: 677 = 00000000000000000000001010100101
681-->1362-->2724-->5448-->10896-->21792-->43584-->21633-->43266--
>20997-->41994-->18453-->36906-->8277-->16554-->33108-->: 16-cycle
state: 681 = 00000000000000000000001010101001
1041-->2082-->4164-->8328-->16656-->33312-->1089-->2178-->4356-->8712--
>17424-->34848-->4161-->8322-->16644-->33288-->: 16-cycle
state: 1041 = 00000000000000000000010000010001
1045-->2090-->4180-->8360-->16720-->33440-->1345-->2690-->5380-->10760-
->21520-->43040-->20545-->41090-->16645-->33290-->: 16-cycle
state: 1045 = 00000000000000000000010000010101
1057-->2114-->4228-->8456-->16912-->33824-->2113-->4226-->8452-->16904-
->33808-->2081-->4162-->8324-->16648-->33296-->: 16-cycle
state: 1057 = 00000000000000000000010000100001
1061-->2122-->4244-->8488-->16976-->33952-->2369-->4738-->9476-->18952-
->37904-->10273-->20546-->41092-->16649-->33298-->: 16-cycle
state: 1061 = 00000000000000000000010000100101
1065-->2130-->4260-->8520-->17040-->34080-->2625-->5250-->10500--
>21000-->42000-->18465-->36930-->8325-->16650-->33300-->: 16-cycle
state: 1065 = 00000000000000000000010000101001
1093-->2186-->4372-->8744-->17488-->34976-->4417-->8834-->17668--
>35336-->5137-->10274-->20548-->41096-->16657-->33314-->: 16-cycle
state: 1093 = 00000000000000000000010001000101
1097-->2194-->4388-->8776-->17552-->35104-->4673-->9346-->18692--
>37384-->9233-->18466-->36932-->8329-->16658-->33316-->: 16-cycle
state: 1097 = 00000000000000000000010001001001
1105-->2210-->4420-->8840-->17680-->35360-->5185-->10370-->20740--
>41480-->17425-->34850-->4165-->8330-->16660-->33320-->: 16-cycle
state: 1105 = 00000000000000000000010001010001
1109-->2218-->4436-->8872-->17744-->35488-->5441-->10882-->21764--
>43528-->21521-->43042-->20549-->41098-->16661-->33322-->: 16-cycle
state: 1109 = 00000000000000000000010001010101

```
1157-->2314-->4628-->9256-->18512-->37024-->8513-->17026-->34052--
>2569-->5138-->10276-->20552-->41104-->16673-->33346-->: 16-cycle
state: 1157 = 00000000000000000000010010000101
1161-->2322-->4644-->9288-->18576-->37152-->8769-->17538-->35076--
>4617-->9234-->18468-->36936-->8337-->16674-->33348-->: 16-cycle
state: 1161 = 00000000000000000000010010001001
1169-->2338-->4676-->9352-->18704-->37408-->9281-->18562-->37124--
>8713-->17426-->34852-->4169-->8338-->16676-->33352-->: 16-cycle
state: 1169 = 00000000000000000000010010010001
1173-->2346-->4692-->9384-->18768-->37536-->9537-->19074-->38148--
>10761-->21522-->43044-->20553-->41106-->16677-->33354-->: 16-cycle
state: 1173 = 00000000000000000000010010010101
1185-->2370-->4740-->9480-->18960-->37920-->10305-->20610-->41220--
>16905-->33810-->2085-->4170-->8340-->16680-->33360-->: 16-cycle
state: 1185 = 00000000000000000000010010100001
1189-->2378-->4756-->9512-->19024-->38048-->10561-->21122-->42244--
>18953-->37906-->10277-->20554-->41108-->16681-->33362-->: 16-cycle
state: 1189 = 00000000000000000000010010100101
1193-->2386-->4772-->9544-->19088-->38176-->10817-->21634-->43268--
>21001-->42002-->18469-->36938-->8341-->16682-->33364-->: 16-cycle
state: 1193 = 00000000000000000000010010101001
1285-->2570-->5140-->10280-->20560-->41120-->16705-->33410-->1285--
>2570-->5140-->10280-->20560-->41120-->16705-->33410-->: 8-cycle
state: 1285 = 00000000000000000000010100000101
1289-->2578-->5156-->10312-->20624-->41248-->16961-->33922-->2309--
>4618-->9236-->18472-->36944-->8353-->16706-->33412-->: 16-cycle
state: 1289 = 00000000000000000000010100001001
1297-->2594-->5188-->10376-->20752-->41504-->17473-->34946-->4357--
>8714-->17428-->34856-->4177-->8354-->16708-->33416-->: 16-cycle
state: 1297 = 00000000000000000000010100010001
1301-->2602-->5204-->10408-->20816-->41632-->17729-->35458-->5381--
>10762-->21524-->43048-->20561-->41122-->16709-->33418-->: 16-cycle
state: 1301 = 00000000000000000000010100010101
1313-->2626-->5252-->10504-->21008-->42016-->18497-->36994-->8453--
>16906-->33812-->2089-->4178-->8356-->16712-->33424-->: 16-cycle
state: 1313 = 00000000000000000000010100100001
1317-->2634-->5268-->10536-->21072-->42144-->18753-->37506-->9477--
>18954-->37908-->10281-->20562-->41124-->16713-->33426-->: 16-cycle
state: 1317 = 00000000000000000000010100100101
1321-->2642-->5284-->10568-->21136-->42272-->19009-->38018-->10501--
>21002-->42004-->18473-->36946-->8357-->16714-->33428-->: 16-cycle
state: 1321 = 00000000000000000000010100101001
1349-->2698-->5396-->10792-->21584-->43168-->20801-->41602-->17669--
>35338-->5141-->10282-->20564-->41128-->16721-->33442-->: 16-cycle
state: 1349 = 00000000000000000000010101000101
1353-->2706-->5412-->10824-->21648-->43296-->21057-->42114-->18693--
>37386-->9237-->18474-->36948-->8361-->16722-->33444-->: 16-cycle
state: 1353 = 00000000000000000000010101001001
1361-->2722-->5444-->10888-->21776-->43552-->21569-->43138-->20741--
>41482-->17429-->34858-->4181-->8362-->16724-->33448-->: 16-cycle
state: 1361 = 00000000000000000000010101010001
1365-->2730-->5460-->10920-->21840-->43680-->21825-->43650-->21765--
>43530-->21525-->43050-->20565-->41130-->16725-->33450-->: 16-cycle
state: 1365 = 00000000000000000000010101010101
2117-->4234-->8468-->16936-->33872-->2209-->4418-->8836-->17672--
>35344-->5153-->10306-->20612-->41224-->16913-->33826-->: 16-cycle
state: 2117 = 00000000000000000000100001000101
```

2121-->4242-->8484-->16968-->33936-->2337-->4674-->9348-->18696--
>37392-->9249-->18498-->36996-->8457-->16914-->33828-->: 16-cycle
state: 2121 = 00000000000000000000100001001001
2129-->4258-->8516-->17032-->34064-->2593-->5186-->10372-->20744--
>41488-->17441-->34882-->4229-->8458-->16916-->33832-->: 16-cycle
state: 2129 = 00000000000000000000100001010001
2133-->4266-->8532-->17064-->34128-->2721-->5442-->10884-->21768--
>43536-->21537-->43074-->20613-->41226-->16917-->33834-->: 16-cycle
state: 2133 = 00000000000000000000100001010101
2181-->4362-->8724-->17448-->34896-->4257-->8514-->17028-->34056--
>2577-->5154-->10308-->20616-->41232-->16929-->33858-->: 16-cycle
state: 2181 = 00000000000000000000100010000101
2185-->4370-->8740-->17480-->34960-->4385-->8770-->17540-->35080--
>4625-->9250-->18500-->37000-->8465-->16930-->33860-->: 16-cycle
state: 2185 = 00000000000000000000100010001001
2193-->4386-->8772-->17544-->35088-->4641-->9282-->18564-->37128--
>8721-->17442-->34884-->4233-->8466-->16932-->33864-->: 16-cycle
state: 2193 = 00000000000000000000100010010001
2197-->4394-->8788-->17576-->35152-->4769-->9538-->19076-->38152--
>10769-->21538-->43076-->20617-->41234-->16933-->33866-->: 16-cycle
state: 2197 = 00000000000000000000100010010101
2213-->4426-->8852-->17704-->35408-->5281-->10562-->21124-->42248--
>18961-->37922-->10309-->20618-->41236-->16937-->33874-->: 16-cycle
state: 2213 = 00000000000000000000100010100101
2217-->4434-->8868-->17736-->35472-->5409-->10818-->21636-->43272--
>21009-->42018-->18501-->37002-->8469-->16938-->33876-->: 16-cycle
state: 2217 = 00000000000000000000100010101001
2313-->4626-->9252-->18504-->37008-->8481-->16962-->33924-->2313--
>4626-->9252-->18504-->37008-->8481-->16962-->33924-->: 8-cycle
state: 2313 = 00000000000000000000100100001001
2321-->4642-->9284-->18568-->37136-->8737-->17474-->34948-->4361--
>8722-->17444-->34888-->4241-->8482-->16964-->33928-->: 16-cycle
state: 2321 = 00000000000000000000100100010001
2325-->4650-->9300-->18600-->37200-->8865-->17730-->35460-->5385--
>10770-->21540-->43080-->20625-->41250-->16965-->33930-->: 16-cycle
state: 2325 = 00000000000000000000100100010101
2341-->4682-->9364-->18728-->37456-->9377-->18754-->37508-->9481--
>18962-->37924-->10313-->20626-->41252-->16969-->33938-->: 16-cycle
state: 2341 = 00000000000000000000100100100101
2345-->4690-->9380-->18760-->37520-->9505-->19010-->38020-->10505--
>21010-->42020-->18505-->37010-->8485-->16970-->33940-->: 16-cycle
state: 2345 = 00000000000000000000100100101001
2373-->4746-->9492-->18984-->37968-->10401-->20802-->41604-->17673--
>35346-->5157-->10314-->20628-->41256-->16977-->33954-->: 16-cycle
state: 2373 = 00000000000000000000100101000101
2377-->4754-->9508-->19016-->38032-->10529-->21058-->42116-->18697--
>37394-->9253-->18506-->37012-->8489-->16978-->33956-->: 16-cycle
state: 2377 = 00000000000000000000100101001001
2385-->4770-->9540-->19080-->38160-->10785-->21570-->43140-->20745--
>41490-->17445-->34890-->4245-->8490-->16980-->33960-->: 16-cycle
state: 2385 = 00000000000000000000100101010001
2389-->4778-->9556-->19112-->38224-->10913-->21826-->43652-->21769--
>43538-->21541-->43082-->20629-->41258-->16981-->33962-->: 16-cycle
state: 2389 = 00000000000000000000100101010101
2581-->5162-->10324-->20648-->41296-->17057-->34114-->2693-->5386--
>10772-->21544-->43088-->20641-->41282-->17029-->34058-->: 16-cycle
state: 2581 = 00000000000000000000101000010101

```
2597-->5194-->10388-->20776-->41552-->17569-->35138-->4741-->9482--
>18964-->37928-->10321-->20642-->41284-->17033-->34066-->: 16-cycle
state: 2597 = 0000000000000000000101000100101
2601-->5202-->10404-->20808-->41616-->17697-->35394-->5253-->10506--
>21012-->42024-->18513-->37026-->8517-->17034-->34068-->: 16-cycle
state: 2601 = 0000000000000000000101000101001
2629-->5258-->10516-->21032-->42064-->18593-->37186-->8837-->17674--
>35348-->5161-->10322-->20644-->41288-->17041-->34082-->: 16-cycle
state: 2629 = 0000000000000000000101001000101
2633-->5266-->10532-->21064-->42128-->18721-->37442-->9349-->18698--
>37396-->9257-->18514-->37028-->8521-->17042-->34084-->: 16-cycle
state: 2633 = 0000000000000000000101001001001
2641-->5282-->10564-->21128-->42256-->18977-->37954-->10373-->20746--
>41492-->17449-->34898-->4261-->8522-->17044-->34088-->: 16-cycle
state: 2641 = 0000000000000000000101001010001
2645-->5290-->10580-->21160-->42320-->19105-->38210-->10885-->21770--
>43540-->21545-->43090-->20645-->41290-->17045-->34090-->: 16-cycle
state: 2645 = 0000000000000000000101001010101
2697-->5394-->10788-->21576-->43152-->20769-->41538-->17541-->35082--
>4629-->9258-->18516-->37032-->8529-->17058-->34116-->: 16-cycle
state: 2697 = 0000000000000000000101010001001
2705-->5410-->10820-->21640-->43280-->21025-->42050-->18565-->37130--
>8725-->17450-->34900-->4265-->8530-->17060-->34120-->: 16-cycle
state: 2705 = 0000000000000000000101010010001
2709-->5418-->10836-->21672-->43344-->21153-->42306-->19077-->38154--
>10773-->21546-->43092-->20649-->41298-->17061-->34122-->: 16-cycle
state: 2709 = 0000000000000000000101010010101
2725-->5450-->10900-->21800-->43600-->21665-->43330-->21125-->42250--
>18965-->37930-->10325-->20650-->41300-->17065-->34130-->: 16-cycle
state: 2725 = 0000000000000000000101010100101
2729-->5458-->10916-->21832-->43664-->21793-->43586-->21637-->43274--
>21013-->42026-->18517-->37034-->8533-->17066-->34132-->: 16-cycle
state: 2729 = 0000000000000000000101010101001
4369-->8738-->17476-->34952-->4369-->8738-->17476-->34952-->4369--
>8738-->17476-->34952-->4369-->8738-->17476-->34952-->: 4-cycle
state: 4369 = 0000000000000000001000100010001
4373-->8746-->17492-->34984-->4433-->8866-->17732-->35464-->5393--
>10786-->21572-->43144-->20753-->41506-->17477-->34954-->: 16-cycle
state: 4373 = 0000000000000000001000100010101
4389-->8778-->17556-->35112-->4689-->9378-->18756-->37512-->9489--
>18978-->37956-->10377-->20754-->41508-->17481-->34962-->: 16-cycle
state: 4389 = 0000000000000000001000100100101
4393-->8786-->17572-->35144-->4753-->9506-->19012-->38024-->10513--
>21026-->42052-->18569-->37138-->8741-->17482-->34964-->: 16-cycle
state: 4393 = 0000000000000000001000100101001
4421-->8842-->17684-->35368-->5201-->10402-->20804-->41608-->17681--
>35362-->5189-->10378-->20756-->41512-->17489-->34978-->: 16-cycle
state: 4421 = 0000000000000000001000101000101
4425-->8850-->17700-->35400-->5265-->10530-->21060-->42120-->18705--
>37410-->9285-->18570-->37140-->8745-->17490-->34980-->: 16-cycle
state: 4425 = 0000000000000000001000101001001
4437-->8874-->17748-->35496-->5457-->10914-->21828-->43656-->21777--
>43554-->21573-->43146-->20757-->41514-->17493-->34986-->: 16-cycle
state: 4437 = 0000000000000000001000101010101
4645-->9290-->18580-->37160-->8785-->17570-->35140-->4745-->9490--
>18980-->37960-->10385-->20770-->41540-->17545-->35090-->: 16-cycle
state: 4645 = 0000000000000000001001000100101
```

4649-->9298-->18596-->37192-->8849-->17698-->35396-->5257-->10514--
>21028-->42056-->18577-->37154-->8773-->17546-->35092-->: 16-cycle
state: 4649 = 00000000000000000001001000101001
4677-->9354-->18708-->37416-->9297-->18594-->37188-->8841-->17682--
>35364-->5193-->10386-->20772-->41544-->17553-->35106-->: 16-cycle
state: 4677 = 00000000000000000001001001000101
4681-->9362-->18724-->37448-->9361-->18722-->37444-->9353-->18706--
>37412-->9289-->18578-->37156-->8777-->17554-->35108-->: 16-cycle
state: 4681 = 00000000000000000001001001001001
4693-->9386-->18772-->37544-->9553-->19106-->38212-->10889-->21778--
>43556-->21577-->43154-->20773-->41546-->17557-->35114-->: 16-cycle
state: 4693 = 00000000000000000001001001010101
4757-->9514-->19028-->38056-->10577-->21154-->42308-->19081-->38162--
>10789-->21578-->43156-->20777-->41554-->17573-->35146-->: 16-cycle
state: 4757 = 00000000000000000001001010010101
4773-->9546-->19092-->38184-->10833-->21666-->43332-->21129-->42258--
>18981-->37962-->10389-->20778-->41556-->17577-->35154-->: 16-cycle
state: 4773 = 00000000000000000001001010100101
4777-->9554-->19108-->38216-->10897-->21794-->43588-->21641-->43282--
>21029-->42058-->18581-->37162-->8789-->17578-->35156-->: 16-cycle
state: 4777 = 00000000000000000001001010101001
5205-->10410-->20820-->41640-->17745-->35490-->5445-->10890-->21780--
>43560-->21585-->43170-->20805-->41610-->17685-->35370-->: 16-cycle
state: 5205 = 00000000000000000001010001010101
5269-->10538-->21076-->42152-->18769-->37538-->9541-->19082-->38164--
>10793-->21586-->43172-->20809-->41618-->17701-->35402-->: 16-cycle
state: 5269 = 00000000000000000001010010010101
5285-->10570-->21140-->42280-->19025-->38050-->10565-->21130-->42260--
>18985-->37970-->10405-->20810-->41620-->17705-->35410-->: 16-cycle
state: 5285 = 00000000000000000001010010100101
5289-->10578-->21156-->42312-->19089-->38178-->10821-->21642-->43284--
>21033-->42066-->18597-->37194-->8853-->17706-->35412-->: 16-cycle
state: 5289 = 00000000000000000001010010101001
5397-->10794-->21588-->43176-->20817-->41634-->17733-->35466-->5397--
>10794-->21588-->43176-->20817-->41634-->17733-->35466-->: 8-cycle
state: 5397 = 00000000000000000001010100010101
5413-->10826-->21652-->43304-->21073-->42146-->18757-->37514-->9493--
>18986-->37972-->10409-->20818-->41636-->17737-->35474-->: 16-cycle
state: 5413 = 00000000000000000001010100100101
5417-->10834-->21668-->43336-->21137-->42274-->19013-->38026-->10517--
>21034-->42068-->18601-->37202-->8869-->17738-->35476-->: 16-cycle
state: 5417 = 00000000000000000001010100101001
5449-->10898-->21796-->43592-->21649-->43298-->21061-->42122-->18709--
>37418-->9301-->18602-->37204-->8873-->17746-->35492-->: 16-cycle
state: 5449 = 00000000000000000001010101001001
5461-->10922-->21844-->43688-->21841-->43682-->21829-->43658-->21781--
>43562-->21589-->43178-->20821-->41642-->17749-->35498-->: 16-cycle
state: 5461 = 00000000000000000001010101010101
9365-->18730-->37460-->9385-->18770-->37540-->9545-->19090-->38180--
>10825-->21650-->43300-->21065-->42130-->18725-->37450-->: 16-cycle
state: 9365 = 00000000000000000010010010010101
9381-->18762-->37524-->9513-->19026-->38052-->10569-->21138-->42276--
>19017-->38034-->10533-->21066-->42132-->18729-->37458-->: 16-cycle
state: 9381 = 00000000000000000010010010100101
9509-->19018-->38036-->10537-->21074-->42148-->18761-->37522-->9509--
>19018-->38036-->10537-->21074-->42148-->18761-->37522-->: 8-cycle
state: 9509 = 00000000000000000010010100100101

```
9557-->19114-->38228-->10921-->21842-->43684-->21833-->43666-->21797--
>43594-->21653-->43306-->21077-->42154-->18773-->37546-->: 16-cycle
state: 9557 = 00000000000000000010010101010101
10581-->21162-->42324-->19113-->38226-->10917-->21834-->43668-->21801--
>43602-->21669-->43338-->21141-->42282-->19029-->38058-->: 16-cycle
state: 10581 = 00000000000000000010100101010101
10837-->21674-->43348-->21161-->42322-->19109-->38218-->10901-->21802--
>43604-->21673-->43346-->21157-->42314-->19093-->38186-->: 16-cycle
state: 10837 = 00000000000000000010101001010101
21845-->43690-->21845-->43690-->21845-->43690-->21845-->43690-->21845--
>43690-->21845-->43690-->21845-->43690-->21845-->43690-->: 2-cycle
state: 21845 = 00000000000000000101010101010101
```

**Photo 1. Hardware system showing the neuron board (detail in Photo 2), and the interconnection.**

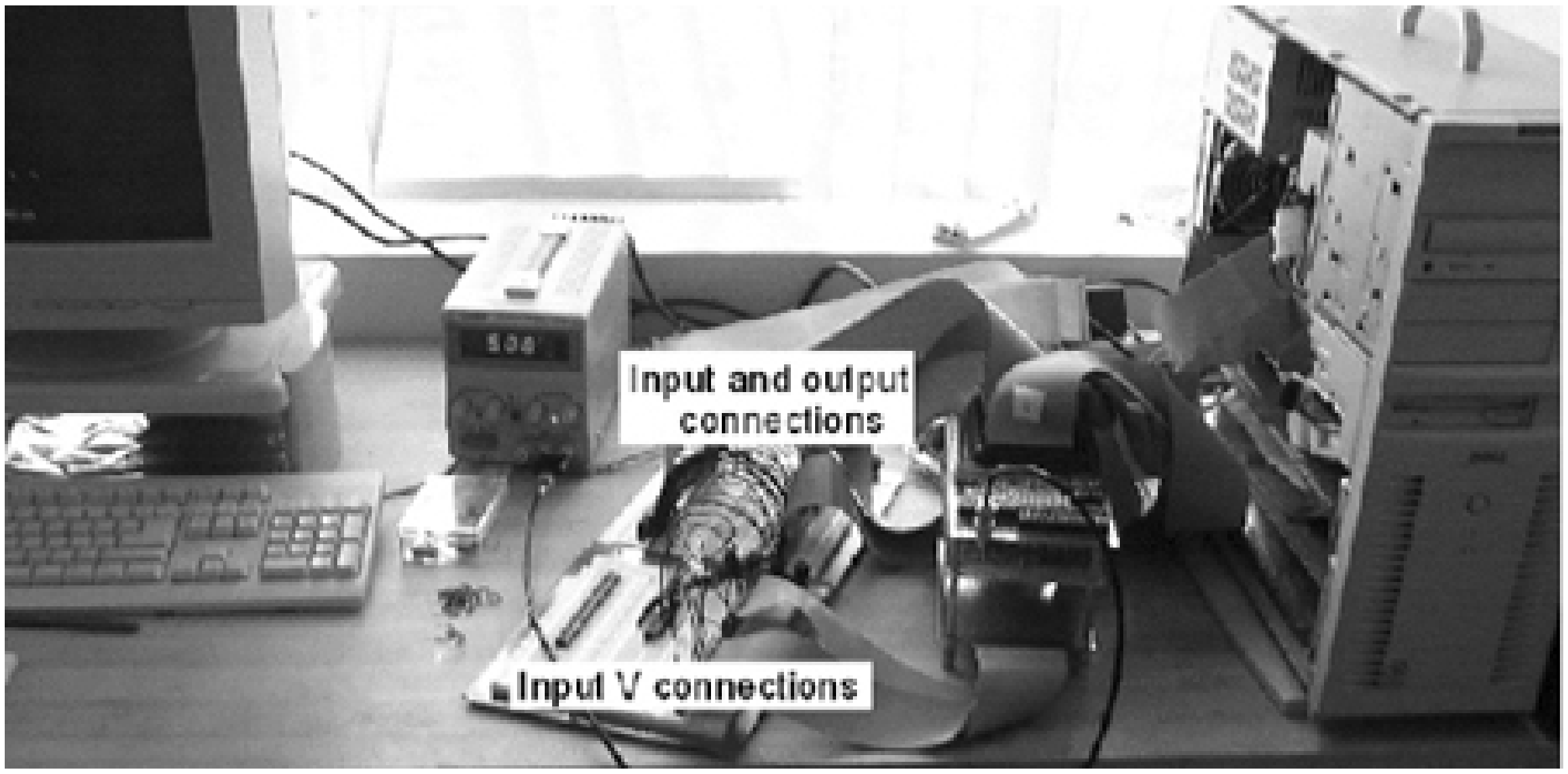

**Photo 2. Neurocomputer with headers, drivers and LEDs.**

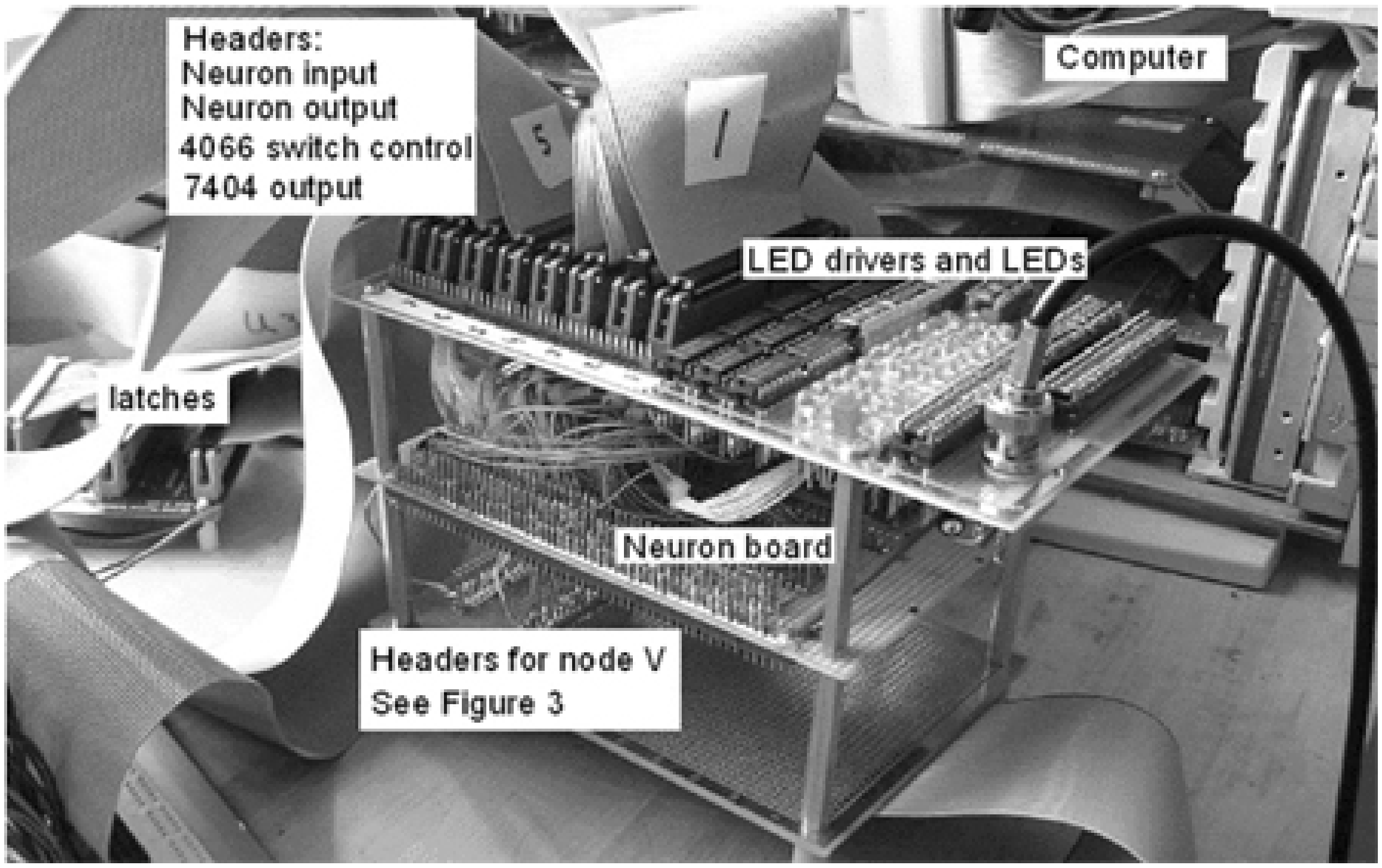

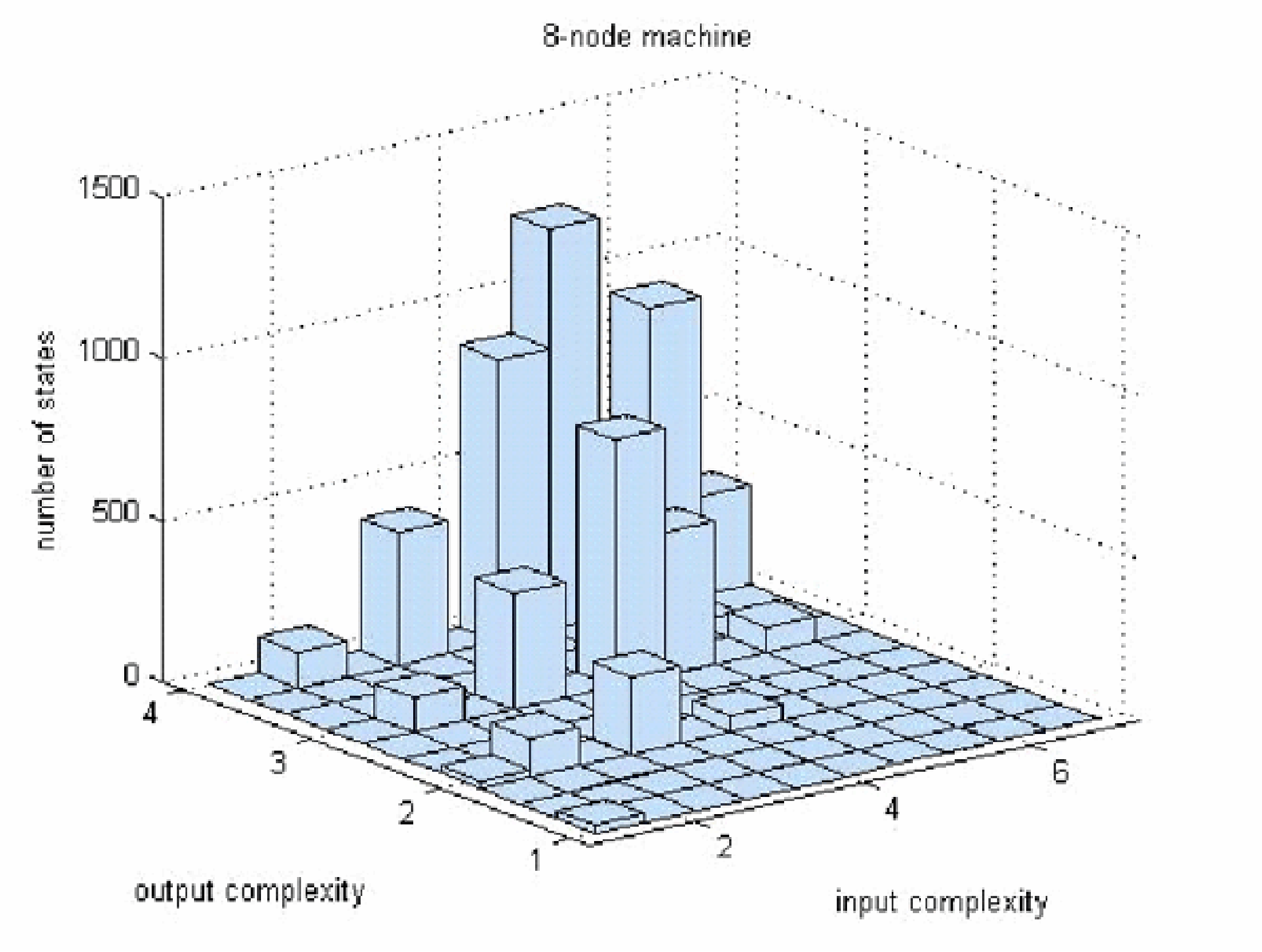

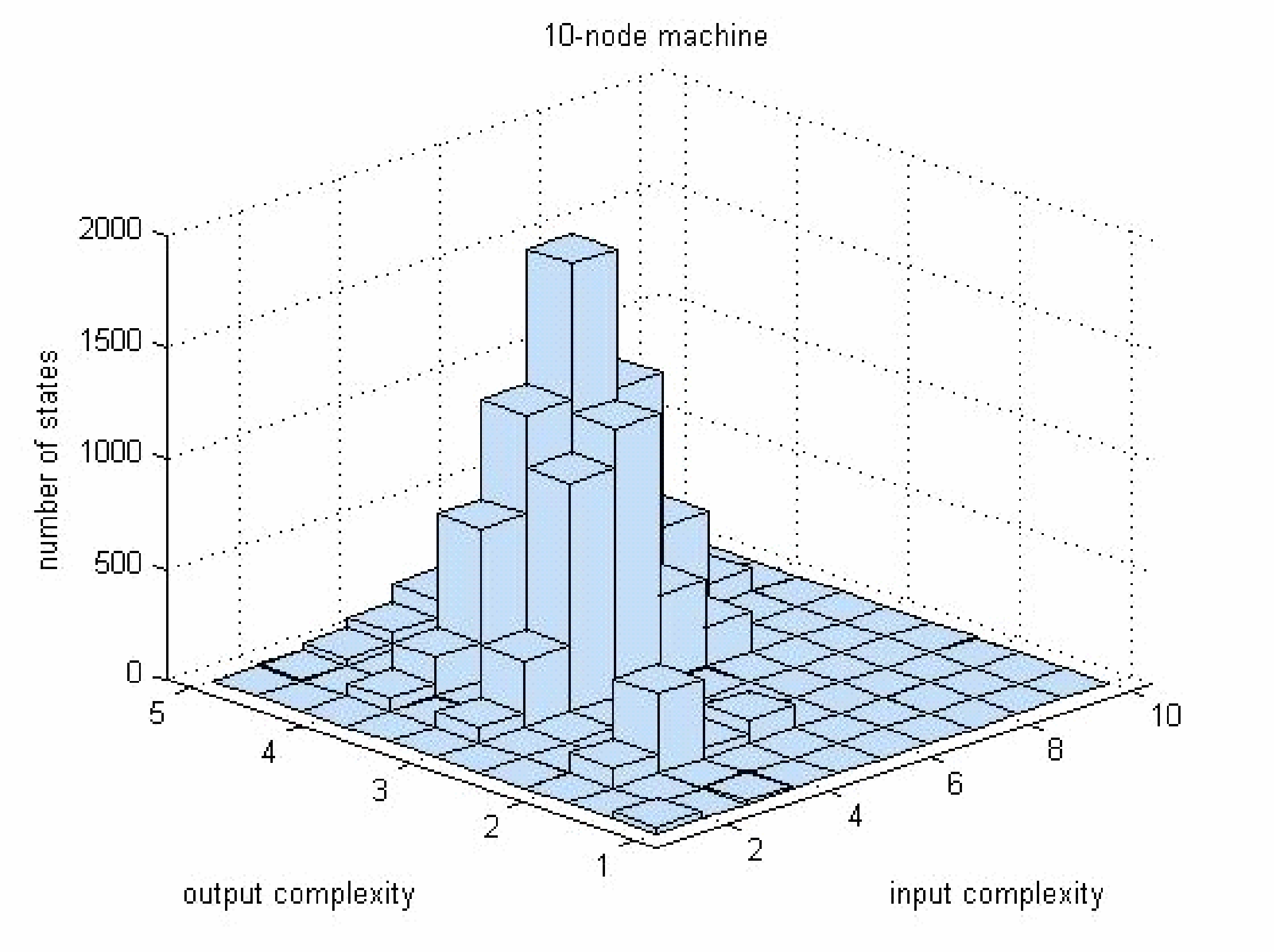

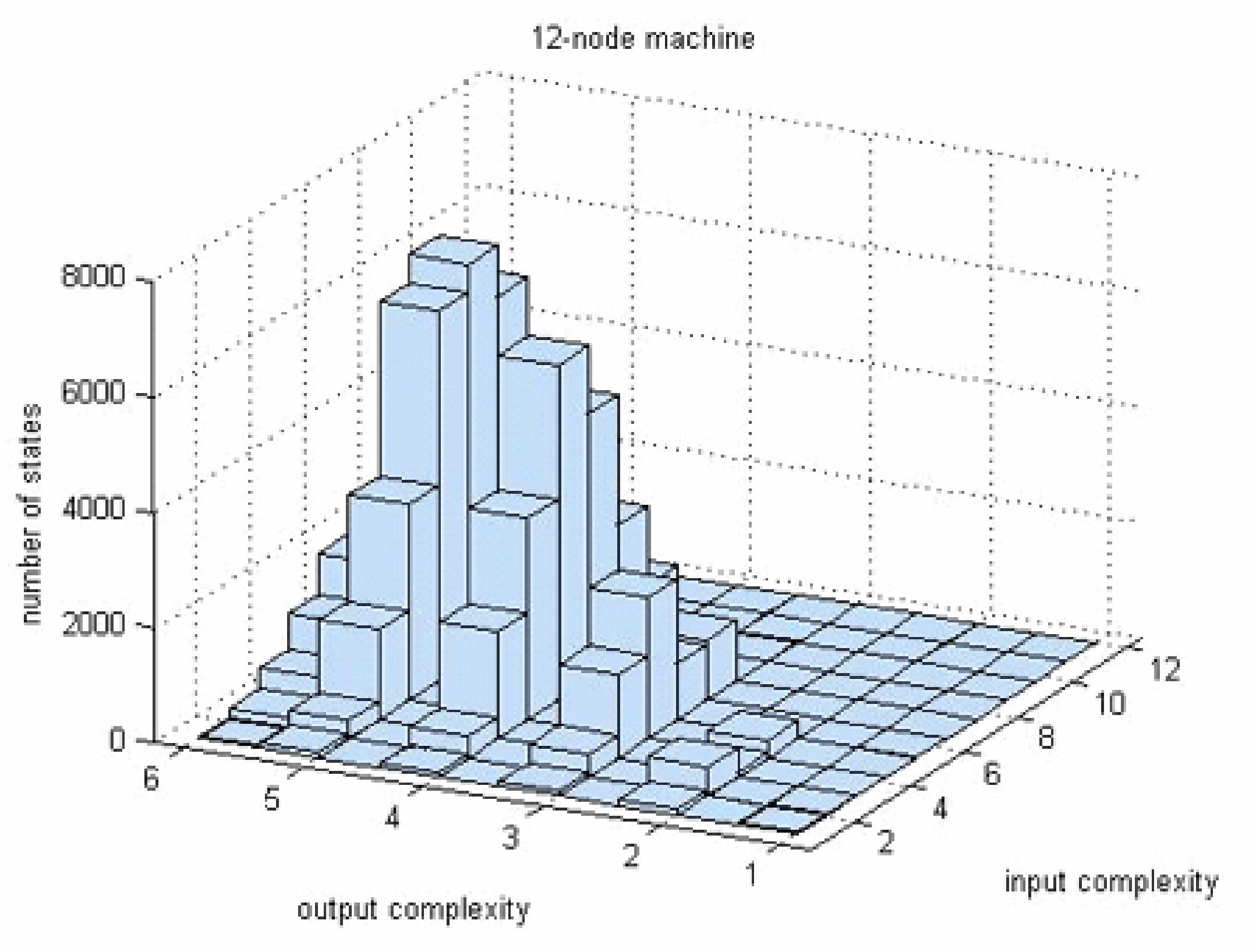

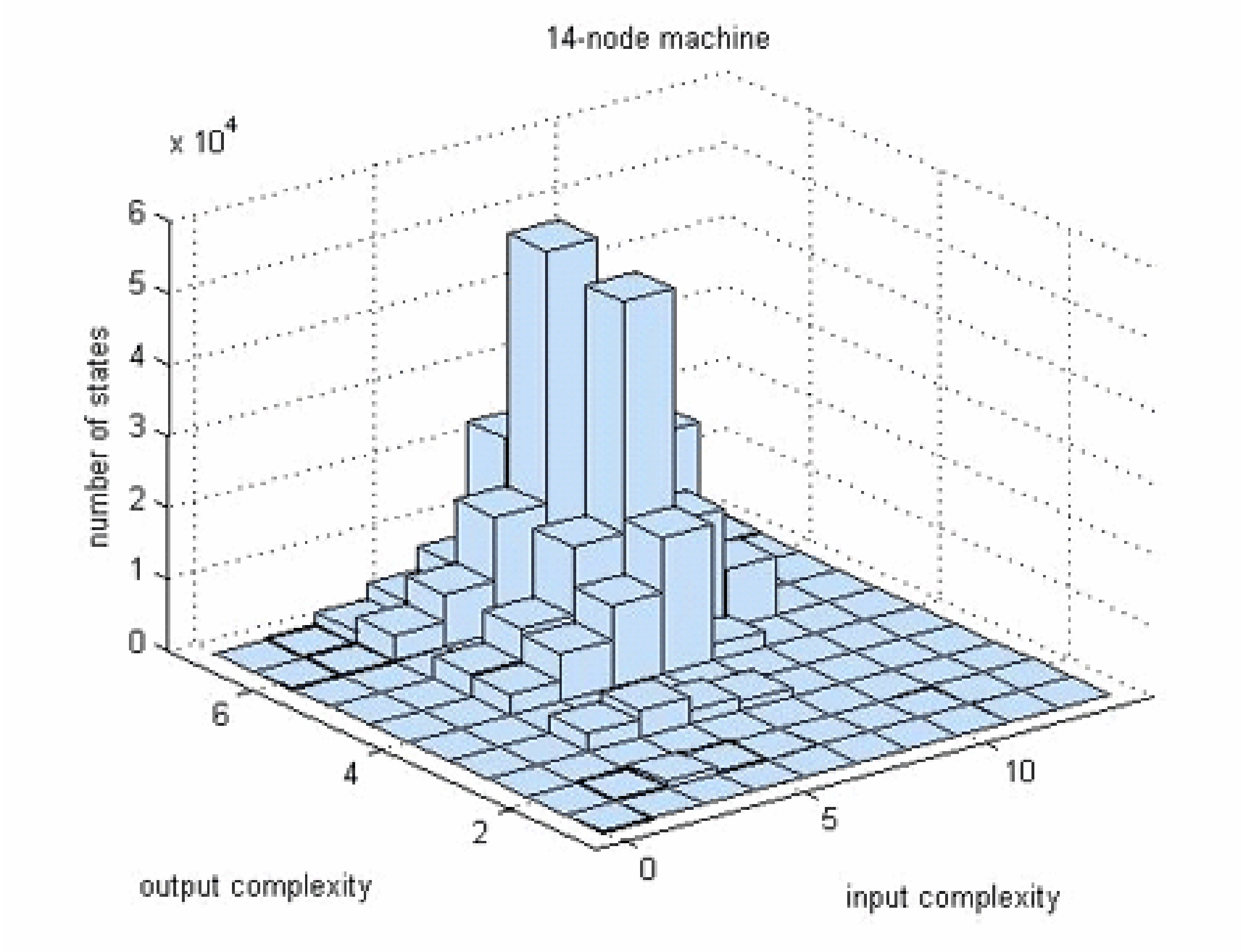

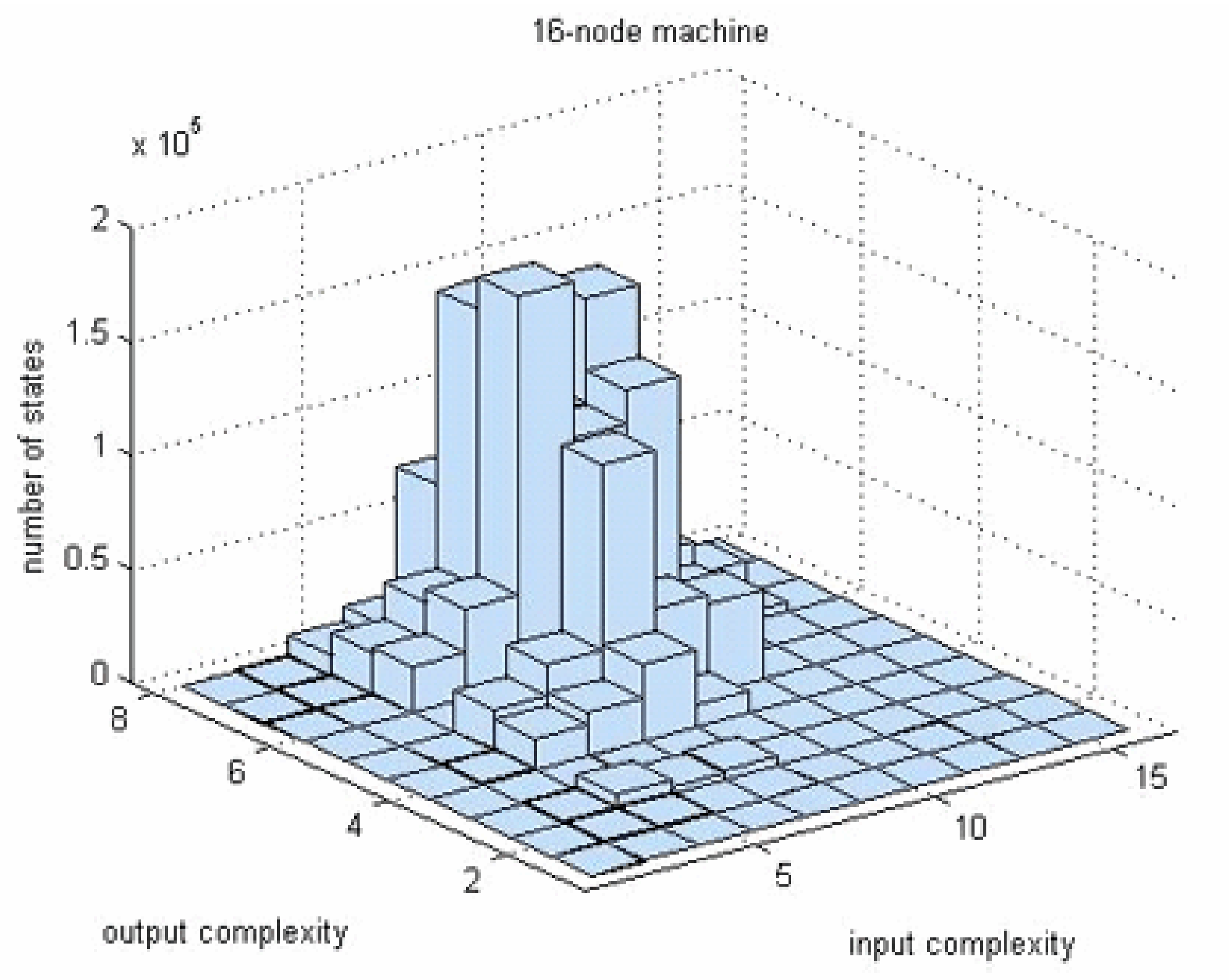